\documentclass[10pt,twocolumn,letterpaper]{article}

\usepackage{wacv}              

\usepackage{color}
\definecolor{igreen}{rgb}{0.0, 0.56, 0.0}
\usepackage{xcolor, colortbl}
\colorlet{gred}{-red!75!green!65!}
\colorlet{mamber}{-red!75!green!15!blue!50!}
\colorlet{grown}{-red!75!blue!20!green}
\colorlet{bled}{-red!85!blue!40!green!45!}
\colorlet{waters}{cyan!25} 
\colorlet{water}{cyan!25!green!20!} 
\definecolor{grin}{HTML}{00F9DE}
\usepackage{rotating}

\usepackage{tikz}
\usepackage{tikz-3dplot}
\usepackage{tikz-cd}
\usetikzlibrary{
arrows,
arrows.meta,
automata,
backgrounds,
calc,
decorations,
decorations.pathmorphing,
decorations.pathreplacing,
decorations.fractals,
external,
fit,
matrix,
petri,
positioning,
shadows,
shapes,
shapes.multipart,
topaths,
intersections
}

\definecolor{wacvblue}{rgb}{0.21,0.49,0.74}
\usepackage[pagebackref,breaklinks,colorlinks,allcolors=wacvblue]{hyperref}

\usepackage{algorithm, algorithmic}
\usepackage{tcolorbox}
\usepackage{setspace}

\def\wacvPaperID{727} 
\def\confName{WACV}
\def\confYear{2026}

\def\model{{\fontfamily{phv}\fontseries{mc}\selectfont brat }}
\def\modelS{{\fontfamily{phv}\fontseries{mc}\selectfont brat}}

\def\bim{BIMCV-R }
\def\dset{MSKBrain}
\def\qformer{Q-Former}

\title{{\fontfamily{phv}\fontseries{mc}\selectfont \colorbox{green}{brat}:} Aligned Multi-View Embeddings for Brain MRI Analysis}

\author{
\textbf{Maxime Kayser}$^{1,2}$\thanks{Corresponding author: \texttt{kayserm1@mskcc.org}} \hspace{1.5em}
\textbf{Maksim Gridnev}$^{1}$ \hspace{1.5em}
\textbf{Wanting Wang}$^{1,3}$ \\
\textbf{Max Bain}$^{2}$ \hspace{2em}
\textbf{Aneesh Rangnekar}$^{1}$ \hspace{2em}
\textbf{Avijit Chatterjee}$^{1}$ \\
\textbf{Aleksandr Petrov}$^{1}$ \hspace{2.5em}
\textbf{Harini Veeraraghavan}$^{1}$ \hspace{2.5em}
\textbf{Nathaniel C. Swinburne}$^{1}$ \\
\\
$^{1}$Memorial Sloan Kettering Cancer Center, New York, United States \\
$^{2}$University of Oxford, United Kingdom \\
$^{3}$London School of Economics, United Kingdom
}

\begin{document}
\maketitle
\begin{abstract}
We present \model (\textbf{b}rain \textbf{r}eport \textbf{a}lignment \textbf{t}ransformer), a multi-view representation learning framework for brain magnetic resonance imaging (MRI) trained on MRIs paired with clinical reports. Brain MRIs present unique challenges due to the presence of numerous, highly varied, and often subtle abnormalities that are localized to a few slices within a 3D volume. To address these challenges, we introduce a brain MRI dataset $10\times$ larger than existing ones, containing approximately 80,000 3D scans with corresponding radiology reports, and propose a multi-view pre-training approach inspired by advances in document retrieval. We develop an implicit query-feature matching mechanism and adopt concepts from quality-diversity to obtain multi-view embeddings of MRIs that are aligned with the clinical features given by report sentences. We evaluate our approach across multiple vision-language and vision tasks, demonstrating substantial performance improvements. The \model foundation models are publicly released\footnote{\url{https://github.com/maximek3/brat}}.
\end{abstract}

\section{Introduction}
\label{sec:intro}

\begin{figure}[t]
  \centering
   \includegraphics[width=1\linewidth]{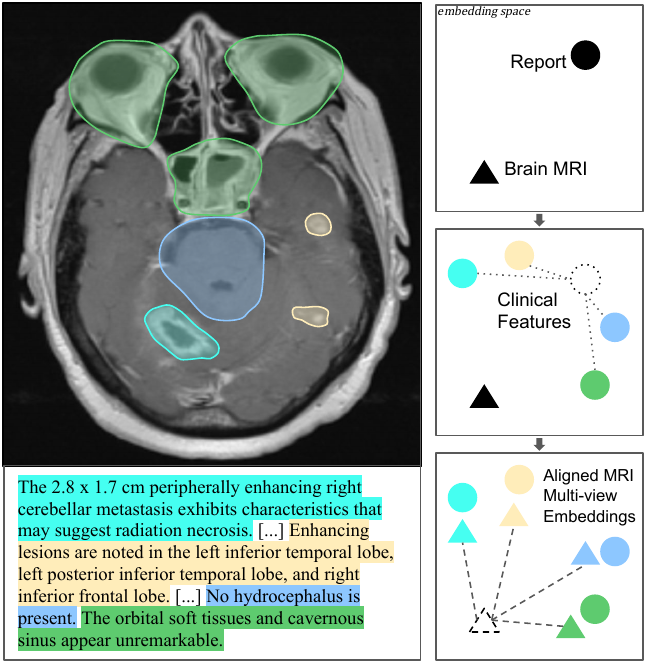}
   \caption{\emph{(Left)} Brain MRI reports contain rich and diverse information relating to different features and regions of the brain. Report sentences are associated with visual features via their colours. The report was cut off ([...]) to only contain findings visible on this 2D slice. \emph{(Right)} By drawing parallels to multi-vector retrieval~\citet{zhang2022multi}, we align multi-view embeddings of the MRI with clinical features given in the reports. Multi-view embeddings can attend across the volume, reflecting that clinical features may correspond to more than one spatial region of the scan. }
   \label{fig:moti}
   \vspace{-0.2cm}
\end{figure}

MRIs are standard-of-care imaging performed for diagnosing and managing neurodegenerative diseases and cancers occurring in the brain. Despite their clinical importance, developing effective AI models for analyzing 3D brain MRI present significant challenges due to the complex anatomy and the limited numbers of well-annotated datasets.

Modern vision-language models (VLM) and vision-language pre-training (VLP) have shown the capability to generate descriptive free-text summarizing medical images~\citep{yang2024advancing} and produce strong vision backbones~\citep{huang_gloria_2021}. However, the scarcity of large-scale medical 3D image-text datasets and the limited generalization of conventional vision-language (VL) methods designed for 2D images have hindered the development of effective VL approaches for 3D imaging modalities. Clinical reports for cross-sectional scans such as CT and MRI often contain rich, diverse information, with multiple sections addressing different aspects of the scan (see Figure~\ref{fig:moti}).
Existing approaches that aim to learn joint representations of medical images and paired text often overlook the complexity and length of these reports. A common approach is to adopt architectures used in natural image captioning, consisting of single sentence descriptions represented via a single embedding. Such an approach fails to fully leverage the richness of clinical narratives as a learning signal to extract diverse image representations~\cite{kovaleva2019revealing, clark-etal-2019-bert}. 

\begin{figure*}
\centering
  \includegraphics[width=1\linewidth]{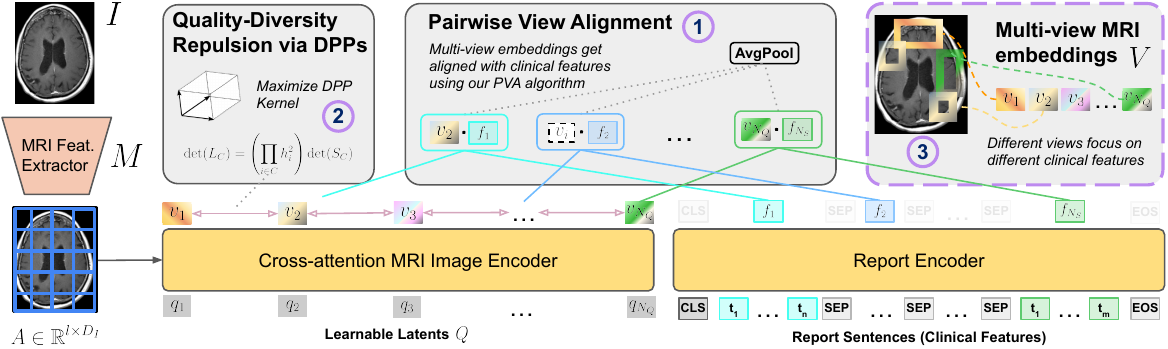}
  \caption{Our \model framework. Our Pairwise View Alignment (PVA) algorithm (described in Section~\ref{sec:pva}) and quality-diversity via Determinental Point Processes (DPPs) (described in Section~\ref{sec:dpp}) lead to clinically aligned multi-view embeddings of the MRI.}
  \label{fig:brat}
\end{figure*}

We address these limitations by introducing \dset, the largest existing multimodal brain MRI dataset with approximately 80,000 3D MRIs and paired reports, and propose a new VLP framework called \model (\textbf{b}rain \textbf{r}eport \textbf{a}lignment \textbf{t}ransformer). 

\model is based on the premise that, similar to documents, clinical reports can consist of extensive text describing a wide range of findings. Multi-view embedding approaches have been used in document retrieval to represent the diversity of information in a document, which can then be mapped to sub-units of the user query~\citep{khattab2020colbert}. In our work, we use such learnable multi-view embeddings to represent the diversity of information in 3D brain MRI, whereas the corresponding reports are subdivided into sentence embeddings or sub-units. By aligning these sub-units with the brain MRI via contrastive learning, we implicitly encourage the multi-view image embeddings to represent the clinical features described by the sentences (see Figure~\ref{fig:moti}). We formulate the multi-view embeddings such that they can represent features that are not restricted to a single spatial location (e.g., brain metastases may appear in multiple sub-regions of the 3D volume). We align the multi-view embeddings with sub-unit sentence embeddings through the proposed Pairwise View Alignment (PVA) matching algorithm and enhance their diversity via a quality diversity (QD) loss based on Determinental Point Processes (DPPs)~\citep{kulesza2012determinantal}. 

Our results show that \model improves accuracy of image-text and text-image retrieval. Vision backbones trained via \model produced higher accuracies on downstream tasks, including report generation, tumour segmentation, and classification of Alzheimer's disease. Our contributions include:
\begin{enumerate}[(i)]
    \item We present the largest ever dataset of 3D radiographs and clinical reports (with 88\% abnormality rate). 
    \item We propose a new multi-view VL representation learning framework tailored for complex 3D medical images by drawing parallels to document retrieval.
    \item We introduce concepts from Quality-Diversity by applying DPPs that encourage diverse and aligned multi-view embeddings.
    \item We pre-train and evaluate state-of-the-art brain MRI foundation models on a wide range of tasks. The weights are publicly released.
\end{enumerate}
\vspace{-0.1cm}
\section{Related Work}

\textbf{Multi-vector document retrieval.} Document retrieval, which involves retrieving documents based on user queries, has seen significant improvements through the use of multi-vector retrieval methods, instead of traditional single-vector approaches. In single-vector retrieval, documents are represented as a single embedding, limiting the ability to capture the diversity of information in documents~\cite{karpukhin2020dense, ni2022large}. In contrast, multi-vector embeddings offer more versatile query-document interactions, which better represent the variety of information in documents. A key work is ColBERT~\cite{khattab2020colbert, santhanam2022colbertv2}, which computes query-document similarity by selecting the most similar document token for each query token and aggregating the similarities across a document. Despite radiology reports often being of document length, these ideas have not yet been applied to image-text datasets. \citet{zhang2022multi} emphasized that documents typically contain multiple semantic units, each relevant to different queries, and proposed using multi-view embeddings to represent these diverse aspects.  Drawing inspiration from their method, we treat brain MRI scans as ``documents'' and report sentences as corresponding ``queries'', to enhance representation and retrieval.

\textbf{Vision-Language Pre-training (VLP).} VLP on large-scale datasets of paired images and captions is an effective way to learn image-representations for both vision~\cite{sun2023eva, radford2021learning, cherti2023reproducible} and VL tasks~\cite{alayrac2022flamingo,chenpali,wanggit}. 
Vision-language datasets occur naturally in medical imaging, as radiologists routinely write reports to describe findings in medical scans. Large-scale public datasets are predominantly available for chest X-rays~\cite{irvin_chexpert_2019, johnson2019mimic, bustos_padchest_2020}, and as such most approaches focus on this domain~\cite{zhang2022contrastive, wu2023medklip, zhang2023knowledge, windsorvision}. Recently, there have been efforts to publish datasets in more advanced imaging modalities, such as lung CTs~\cite{chen2024bimcv}. Some existing models attempt to capture the fine-grained features of medical images by aligning local image feature patches with text tokens~\cite{huang_gloria_2021, wang2022multi}. However, this is limited by the fact that individual text tokens are not necessarily representative of clinical features, and image patches are restricted to a single spatial region in the image. In addition, chest X-rays are 2D images and their reports often only contain 2-3 image-descriptive sentences. 3D scans such as brain MRIs, typically have reports that are several times larger. Recently, the first models have been trained on large 3D lung CTs and report datasets, relying mainly on scale to achieve good results~\cite{yang2024advancing}.

A suite of existing VLMs use learnable latent variables to efficiently compress visual representations~\cite{jaegle2021perceiver,li2023blip}.
BLIP-2~\cite{li2023blip}, for example, employs a Q-Former model that uses ``querying tokens'' as learnable latents that align cross-modal representations. We adopt a similar architecture, with the querying tokens representing the learnable multi-view embeddings that we align to clinical features.

\begin{figure}[t]
  \centering
   \includegraphics[width=1\linewidth]{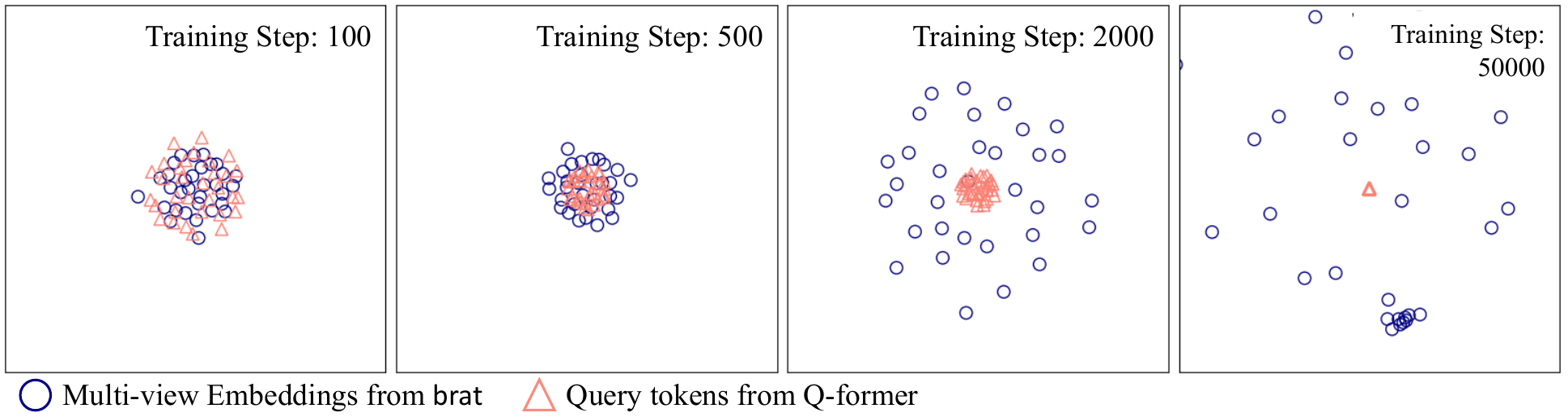}
   \caption{Conventional query tokens collapse into a single representation as training progresses. The multi-view embeddings of \modelS, on the other hand, are diverse and spread out. 
   The plot was obtained by multi-dimensional scaling of 32 query tokens to 2D based on their mean pairwise distances from 32 images.
   }
   \label{fig:diversity}
   \vspace{-0.35cm}
\end{figure}

\vspace{0.1ex}
\textbf{Quality-diversity of learned features.} 
Increasing diversity of features can be useful to avoid informational collapse in self-supervised learning (SSL)~\cite{zbontar2021barlow, bardes2022vicreg} or to capture different aspects of an input, e.g. in document retrieval~\cite{zhang2022multi}. To achieve this, \citet{zhang2022multi} penalize high pairwise similarity between embedding vectors.
Determinantal point processes (DPPs), however, measure of the overall spread of diversity and integrate quality notions seamlessly~\citep{kulesza2012determinantal}. Whilst they have been used for increasing quality and diversity in recommender systems, summarisation, or dataset/batch sampling~\citep{li2021exploring, saran2023streaming}, they have not yet been applied to multi-view representation learning. We show that they are well-suited for this and outperform simple pairwise similarity reduction.

\textbf{Brain MRI analysis.} The scarcity of large-scale brain MRI datasets (and lack of VL datasets) has led researchers to pool smaller public datasets~\cite{munk2024amaes} and focus on SSL tasks (e.g. reversing various image augmentations~\cite{vincent2008extracting,zhang2016colorful,noroozi2016unsupervised}, masked image modeling~\cite{beit2022,he2022masked,tang2022self,jiang2022self}, or contrastive losses~\cite{chen2020simple}). Other efforts aim to generalize across MRI modalities~\cite{konwer2023enhancing, xufeasibility}. Downstream applications often focus on segmentation, notably via the BraTS datasets~\cite{baid2021rsna,moawad2023brain}, on which we evaluate our models. \dset \ is around $10\times$ larger than the biggest available brain MRI dataset (80K vs. 8K MRI sessions). Moreover, it comes with radiology reports, structured labels, and is extremely rich in anomalies. We show that it leads to improved representation learning and downstream performance.
\vspace{-0.05cm}
\section{Methods}

\subsection{The \dset \ Dataset}

We collected a comprehensive dataset of brain MRI scans and their corresponding clinical reports from a cancer center, covering the period from 2012 to 2017. These scans were primarily obtained to monitor brain metastases and tumors in cancer patients, resulting in a dataset rich in positive findings (87.9\% of scans show abnormalities) and representative of a diverse patient population. We also collected extensive demographic data, primary diagnosis, ongoing chemotherapy and radiotherapy treatments, and survival information, which will be considered in future studies. Our dataset includes 77,228 brain MRI image-report pairs from 24,262 unique patients. To develop our model, we performed a patient-wise split of the data into 75,142 examples for training, 945 for development, and 1,141 for the test set. Further details are given in Table~\ref{tab:dataset_summary}.

As the focus of this work is on learning image representations of brain MRIs, we ensured that all report content was visually grounded in the corresponding images. For example, keyword filtering revealed that 94\% of reports make references to prior scans. To efficiently remove these references, as well as information from excluded MRI modalities and protected health information (PHI), we developed a PHI-enabled GPT-4 pipeline. This pipeline simultaneously re-wrote reports to make them visually grounded and extracted structured data for further use. \citet{liu2023exploring} demonstrated that GPT-4 performs well on radiology report processing; and indeed, we found that our pipeline achieved an annotation accuracy of 96\% on a gold standard set of 50 manually annotated reports. Annotating all the reports cost approximately \$1,600, which is significantly lower than the cost of expert annotation. More information on the dataset and processing is provided in Appendix~\ref{app:dset}. 

\begin{table}
  \centering
  \resizebox{\columnwidth}{!}{%
    \begin{tabular}{@{}lc@{}}
      \toprule
      Characteristic & Value \\
      \midrule
      Word Count (Q1, Median, Q3) & 115, 134, 156 \\
      Sentence Count (Q1, Median, Q3) & 7, 9, 11 \\
      Age (Q1, Median, Q3) & 45, 58, 68 \\
      Any Abnormality (\%) & 87.9 \\
      Prior Surgery (\%) & 38.1 \\
      Enhancing Lesions (\%) & 47.6 \\
      Midline Shift (\%) & 5.4 \\
      White Matter Changes (\%) & 43.6 \\
      Pituitary Gland Abnormality (\%) & 2.3 \\
      Hydrocephalus (\%) & 2.6 \\
      Biggest Mass Length (\%) & $<$1cm (17.7), 1-2cm (14.5) \\ & $>$2cm (18.8) \\
      Enhancing Lesion Count (\%) & 1 (27.0), 2-6 (23.0) \\ & 7-15 (1.6), $>$15 (2.3) \\
      \# of Unique Surgeries & 32,428 \\
      \# of Enhancing Lesion Locations & 95,815 \\
      \bottomrule
    \end{tabular}%
  }
  \caption{Brain MRI dataset characteristics. For reference, Conceptual Captions~\cite{sharma2018conceptual} has 10 tokens ($<10$ words) per image.}
  \vspace{-0.1cm}
  \label{tab:dataset_summary}
\end{table}

\subsection{The \model Framework} \label{sec:brat}

Volumetric brain MRIs are visually complex due to the varied appearance of brain tumors, their effects on surrounding tissue, and findings being diverse and spread across the 3D volume. Multi-view embeddings have been shown to improve document representations by capturing different semantic elements within a text~\cite{zhang2022multi}. By drawing parallels between the distinct set of findings in brain MRIs and the distinct semantic units of documents, we hypothesize that multi-view embeddings provide a more flexible, and thus more suitable, representation for brain MRI images. Specifically, we assume that individual sentences in radiology reports correspond to distinct clinical features, and we align the multi-view embeddings to encapsulate these same clinical characteristics (see Figure~\ref{fig:moti}). 

We model this using \modelS, a vision-language contrastive pre-training framework that represents images via aligned multi-view embeddings. We obtain multi-view embeddings by adopting a base architecture similar to \qformer~\cite{li2023blip}, i.e., by having learnable latents to extract multi-view embeddings by cross-attending to localized MRI features (see Fig.~\ref{fig:brat}). A 3D vision model $M$ is used to extract these features from an MRI image $I$, resulting in a set of feature maps $M(I) = A \in \mathbb{R}^{l \times D_I}$ with $l$ feature maps of dimension $D_I$. We used Densenet-121~\citep{Huang_2017_CVPR} as $M$, because it outperformed ViT and Resnet-50 in preliminary experiments. The set of learnable latent tokens $Q = [{q_1, \dots, q_{N_Q}}]$ where $q_l \in \mathbb{R}^{D_Q}$, interact with the image encoder features $A$ to extract a set of image-informed multi-view embeddings $E_I(Q, I) = V$, where $V = [{v_1, \dots, v_{N_Q}}]$ with  $v_i \in \mathbb{R}^{D_V}$. The text encoder $E_R$ takes a brain MRI radiology report 
and returns sentence embeddings that capture the clinical features described in them: $E_R(R) = F$, where $F = [{f_1, \dots, f_{N_S}}]$ with  $f_i \in \mathbb{R}^{D_F}$ representing the $i$-th sentence. We obtain sentence embeddings $f_i$ by averaging all token embeddings of the sentence. As $D_F = D_V = D$, $D$ is used in the rest of the paper for clarity. Implementation details are provided in Appendix~\ref{app:trai9}. 

Existing approaches that use latent variables to extract image features often exhibit embedding collapse, where the learned latents converge into a single representation~\cite{chauhan2024continuous} (an illustration is given in Figure~\ref{fig:diversity}). To ensure that we obtain multi-view embeddings focusing on distinct clinical features, we introduce a two-step approach: (1) Pairwise View Alignment (PVA) to align embeddings with clinically meaningful features, and (2) quality-diversity repulsion using determinantal point processes (DPPs) to encourage diversity in the learned representations.

\paragraph{Pairwise View Alignment} \label{sec:pva}

The PVA algorithm aligns the multi-view embeddings with clinically meaningful features, i.e., sentence embeddings. PVA formulates the alignment as a greedy bipartite matching problem, ensuring each image view is matched to at most one unique sentence embedding. As outlined in Algorithm~\ref{alg:pva}, the computational cost is dominated by sorting the similarity pairs, resulting in a time complexity of $O(N_Q N_S \log(N_Q N_S))$.
The overall image-report similarity, used for the contrastive loss, is then given by the average of all multi-view embeddings and matched clinical feature similarities. 

\begin{algorithm}[h]
    \caption{Pairwise View Alignment (PVA)}
    \label{alg:pva}
    \textbf{Input:} Normalized multi-view embeddings $V \in \mathbb{R}^{N_Q \times D}$ and report features $F \in \mathbb{R}^{N_S \times D}$ \\
    \textbf{Output:} Set of matched pairs $P_{M}$

    \begin{algorithmic}[1]
    \STATE Compute similarity matrix $S \in \mathbb{R}^{N_Q \times N_S}$ via \\ $S \gets V F^T$

    \STATE Create a list $\mathcal{L}$ of all tuples $(i, j, S_{i,j})$ sorted in descending order of similarity $S_{i,j}$

    \STATE Initialize matched pairs $P_M \gets \varnothing$
    \STATE Initialize sets of occupied indices $\mathcal{O}_V \gets \varnothing, \mathcal{O}_F \gets \varnothing$

    \FOR{each $(i, j, \text{sim})$ in $\mathcal{L}$}
        \STATE \COMMENT{Select pair only if both view $i$ and feature $j$ are unmatched}
        \IF{$i \notin \mathcal{O}_V \land j \notin \mathcal{O}_F$}
            \STATE Add $(i,j)$ to matchd pairs: $P_M \gets P_M \cup \{(i,j)\}$
            \STATE Mark indices as occupied: \\ $\mathcal{O}_V \gets \mathcal{O}_V \cup \{i\} \text{ ; } \mathcal{O}_F \gets \mathcal{O}_F \cup \{j\}$
        \ENDIF
        \STATE \textbf{break if} $|P_M| = \min(N_Q, N_S)$
    \ENDFOR
    \end{algorithmic}
\end{algorithm}

\vspace{-0.2cm}
\paragraph{Quality-Diversity via DPPs} \label{sec:dpp} 

\begin{figure}[h]
  \centering
  \includegraphics[width=1\linewidth]{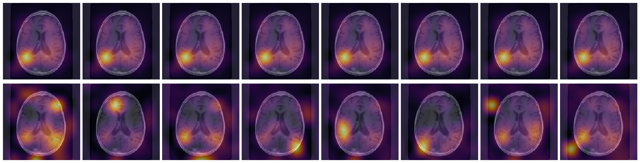}
   \caption{Juxtaposition of 8 query tokens from Q-Former (upper row) and the same 8 tokens from \model (lower row). The collapsed Q-Former queries all attend to the same image regions, whereas the multi-view embeddings of \model focus on distinct features.} 
    \vspace{-0.2cm}
   \label{fig:attn}
\end{figure}

Empirically, we find that PVA alone does not sufficiently encourage diverse features in multi-view embeddings. To address this, we adopt ideas from quality-diversity (QD). The idea behind QD is to have many diverse solutions to tackle a problem from different angles. This fits our problem well, as we want the different multi-view embeddings to focus on different features of the image. We consider as feature diversity the diversity of the attention maps over the image feature maps $A$ of the multi-view embeddings $V$. 

Pairwise dissimilarity measures, e.g., used in~\citet{zhang2022multi}, enforce local repulsion between embeddings but may result in trivial solutions. In contrast, DPPs explicitly model global diversity by maximizing the volume spanned by embeddings collectively, thus avoiding trivial solutions and encouraging each embedding to capture distinct aspects of the feature space~\citep{kulesza2012determinantal}. In addition, DPPs allow the seamless integration of a quality heuristic. Figure~\ref{fig:dpp-moti1} and~\ref{fig:dpp-moti2} in the Appendix illustrate how DPPs promote a more desirable feature diversity than pairwise repulsion. The results in Table~\ref{tab:msk} empirically validate the superiority of DPPs.

DPPs are distributions over subsets of a fixed ground set that attribute higher probability to sets that are diverse. In our case, we want to maximize the probability of our set of multi-view embeddings $V$ under the DPP.
We consider quality-diversity with respect to the cross-attention maps $C = [{c_1, \dots, c_{N_Q}}]$ where $c_j$ contains the flattened attention values from multi-view embedding $v_j$ to the 3D feature maps representing the image. As \emph{quality} of each embedding token we use Shannon entropy of its cross-attention map, denoted as $h_i \in \mathbb{R}^+$: 
\begin{equation} h_i = \mathcal{H}(c_i) = -\sum_{k} c_i(k) \log c_i(k), \end{equation}

where $k$ indexes over spatial positions. 
In this context, we interpret high entropy as high information capacity: it encourages the embedding to capture broader, region-level semantic features rather than collapsing into a trivial, single-voxel representation. This ensures that the diversity term acts on meaningful feature regions rather than disjoint pixels.
The attention maps $c_i$ themselves are considered as the \emph{diversity features}. The DPP kernel matrix $L_{ij}$ can be written as:

\begin{equation}
L_{ij} = h_ic_i^Tc_jh_j.
\end{equation}

The DPP for a selected subset $C'$ is given by:

\begin{equation}
P_L(C') \propto \det(L_{C'}),
\end{equation}

In our case $C=C'$, as we consider repulsion between all image tokens. $\det(L_{C})$ can be decomposed as follows:

\begin{equation}
\det(L_{C}) = \left( \prod_{i \in C} h_i^2 \right) \det(S_{C}),
\end{equation}

where $S_{C}$ is the similarity matrix between all attention maps $c_i$.

The determinant of the kernel matrix $L_{C}$ corresponds to the squared volume of the parallelepiped spanned by the vectors $h_i c_i$ for each $i$ in $C$. By maximizing the product $\prod_{i \in C} h_i^2$, we encourage each embedding token to have high entropy, corresponding to a large magnitude in the feature space. By maximizing the determinant $\det(S_{C})$, where $S_{C}$ captures the pairwise similarities between attention maps, we ensure that the directions $c_i$ are as different as possible, promoting diversity among the tokens. This approach naturally prevents the embeddings from collapsing into a single representation by encouraging both high quality (non-collapsed attention) and diverse (distinct attention patterns) embeddings. 

In practice, we define the DPP loss by taking the negative log-determinant of the kernel matrix $L$

\begin{equation} \mathcal{L}_{\text{DPP}} = -\log \det(L_C + \epsilon I), \end{equation}

where $\epsilon I$ is a small diagonal matrix added for numerical stability.

\paragraph{Loss Calculation}

To obtain the overall image-report similarity, we aggregate the multi-view embedding sentence similarities by mean-averaging:

\begin{equation}
S_{R,I} = S_{I,R} = \frac{1}{|P_M|} \sum_{(i, j) \in P_M} S_{v,f}[i, j]
\end{equation}

As such, we get our contrastive losses as follows:

\begin{equation}
\mathcal{L}^{(I|R)} = -\log\left(\frac{\exp(S_{I,R}/\tau)}{\sum_{k} \exp(S_{I,R_k}/\tau)}\right)
\end{equation}

\begin{equation}
\mathcal{L}^{(R|I)} = -\log\left(\frac{\exp(S_{R,I}/\tau)}{\sum_{m} \exp(S_{R,I_m}/\tau)}\right)
\end{equation}

We also use the same ``Image-grounded Text Generation'' (ITG) loss as in BLIP-2~\cite{li2023blip}, as we found it to help performance. Our final loss is thus given as:

\begin{equation}
\mathcal{L} = \frac{\mathcal{L}^{(I|R)} + \mathcal{L}^{(R|I)}}{2} + \mathcal{L}_{\text{DPP}} + \mathcal{L}_{\text{ITG}}
\end{equation}

\subsection{\model as a Foundation Model} \label{sec:downstream}

Our \model framework provides both a pre-trained vision backbone $M$ and a model $E_I(Q, I)$ for extracting multi-view embeddings. We refer to \modelS-viz for the vision backbone only, and \model for multi-view framework.

\begin{figure}[h]
  \centering
   \includegraphics[width=1\linewidth]{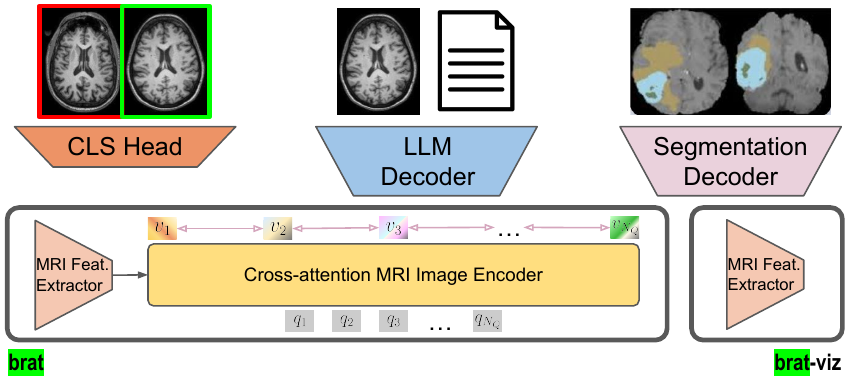}
   \caption{We connect two configurations of \model with various decoders to evaluate our pre-training on downstream tasks.}
   \label{fig:downstream}
\end{figure}

Figure~\ref{fig:downstream} illustrates how the \model weights can be modularized for different downstream tasks. Different task-specific heads, such as an MLP for classification, a language model for report generation, or a segmentation decoder, can appended to either \model or \modelS-viz. Experiments on different such configurations are provided in the next section.  

\vspace{-2ex}
\section{Experiments and Results}

This paper presents a new VLP method for 3D medical scans and document-length reports. In this section, we demonstrate the benefits of \model over other pre-training methods, both in terms of pre-training metrics and on downstream tasks, including tumor and metastases segmentation, Alzheimer's classification, and report generation.

\begin{table*}[ht]
  \centering
  \resizebox{\textwidth}{!}{%
  \begin{tabular}{l*{14}{c}}
    \toprule
    Methods & \multicolumn{7}{c}{Text to Image} & \multicolumn{7}{c}{Image to Text} \\
    \cmidrule(lr){2-8} \cmidrule(lr){9-15}
    & R@1 $\uparrow$ & R@5 $\uparrow$ & R@10 $\uparrow$ & R@5 (F) $\uparrow$ & P@5 (F) $\uparrow$ & MdR $\downarrow$ & MnR $\downarrow$ 
    & R@1 $\uparrow$ & R@5 $\uparrow$ & R@10 $\uparrow$ & R@5 (F) $\uparrow$ & P@5 (F) $\uparrow$ & MdR $\downarrow$ & MnR $\downarrow$ \\
    \midrule
    CLIP            & 0.146    & 0.407    & 0.564    & 0.894    & 0.718    & 8.0     & 35.8   & 0.159    & 0.431    & 0.569    & 0.853    & 0.748    & 8.0     & 37.2 \\
    QFormer         & 0.154    & 0.377    & 0.529    & 0.867    & 0.672    & 10.0    & 32.5   & 0.146    & 0.368    & 0.532    & 0.837    & 0.703    & 9.0     & 34.9 \\
    Colbert         & 0.125    & 0.370    & 0.509    & 0.889    & 0.680    & 10.0    & \textbf{31.9}   & 0.113    & 0.326    & 0.487    & 0.810    & 0.732    & 11.0    & 36.0 \\
    \model w/o QD     & 0.173    & 0.458    & 0.615    & 0.894    & 0.711    & \textbf{6.0}    & 37.3   & 0.171    & 0.449    & 0.606    & 0.875    & 0.745    & 7.0     & \textbf{34.5} \\
    brat w/ PR      & 0.099    & 0.349    & 0.497    & 0.875    & \textbf{0.723}    & 11.0    & 36.6   & 0.109    & 0.328    & 0.478    & 0.818    & 0.696    & 11.0    & 39.2 \\
    QFormer w/ QD   & 0.155        & 0.370        & 0.542        & 0.851        & 0.701        & 10.0       & 34.8        & 0.152        & 0.381        & 0.529        & 0.817        & 0.699        & 10.0       & 33.6 \\
    \model          & \textbf{0.205}    & \textbf{0.493}    & \textbf{0.666}    & \textbf{0.911}    & 0.718    & \textbf{6.0}    & 124.1  & \textbf{0.201}    & \textbf{0.481}    & \textbf{0.645}    & \textbf{0.882}    & \textbf{0.752}    & \textbf{6.0}    & 96.3 \\
    brat vit        & 0.015    & 0.066    & 0.117    & 0.661    & 0.410    & 385.0   & 404.6  & 0.016    & 0.066    & 0.129    & 0.604    & 0.473    & 357.0   & 401.0 \\
    brat resnet     & 0.095    & 0.292    & 0.436    & 0.843    & 0.654    & 13.0    & 109.4  & 0.131    & 0.343    & 0.462    & 0.809    & 0.640    & 12.0    & 62.0 \\
    \bottomrule
  \end{tabular}%
  }
  \caption{Evaluation results for text-to-image and image-to-text retrieval on \dset. For the “$\uparrow$” metrics higher is better and for the “$\downarrow$” metrics lower is better. “R@5 (finding)” and “P@5 (finding)” indicate the recall and precision at 5 for the finding task.}
  \label{tab:msk}
\end{table*}

\begin{table*}[h]
  \centering
  \resizebox{\textwidth}{!}{%
  \begin{tabular}{lcccccccccc}
  \toprule
  Methods & \multicolumn{5}{c}{Text to Image} & \multicolumn{5}{c}{Image to Text} \\
  \cmidrule(lr){2-6} \cmidrule(lr){7-11}
         & R@1 $\uparrow$ & R@5 $\uparrow$ & R@10 $\uparrow$ & MdR $\downarrow$ & MnR $\downarrow$ & R@1 $\uparrow$ & R@5 $\uparrow$ & R@10 $\uparrow$ & MdR $\downarrow$ & MnR $\downarrow$ \\
  \midrule
  CLIP4clip [21]       & 0.003  & 0.015  & 0.022  & 717.0  & 735.9  & 0.003  & 0.008  & 0.015  & 722.0  & 738.7  \\
  3D-MIR [1]           & 0.011  & 0.047  & 0.103  & 121.1  & 152.3  & 0.012  & 0.040  & 0.088  & 134.9  & 162.4  \\
  MedFinder (Resnet-50) & 0.028  & 0.087  & 0.203  & \textbf{68.9}  & 81.3   & 0.029  & 0.088  & 0.197  & 71.2   & \textbf{80.7}  \\
  MedFinder (ViT-base)  & 0.027  & 0.089  & \textbf{0.214}  & 75.4   & \textbf{80.1}  & 0.027  & 0.090  & \textbf{0.203}  & 72.3   & 81.9  \\
  Q-Former             & 0.007  & 0.025  & 0.048  & 223.0  & 371.7  & 0.000  & 0.015  & 0.034  & 225.0  & 365.8  \\
  \model               & \textbf{0.030}  & \textbf{0.109}  & 0.165  & 71.0   & 283.0  & \textbf{0.036}  & \textbf{0.103}  & 0.182  & \textbf{67.0}  & 282.0  \\
  \bottomrule
  \end{tabular}%
  }
  \caption{Evaluation results for text-to-image and image-to-text retrieval on BIMCV-R, a lung CT dataset. The compared results, except Q-Former, are taken from~\citet{chen2024bimcv}. It's unclear how the median ranks happened to be reported as non discrete values.}
  \label{tab:bim}
\end{table*}

\subsection{VLP Performance: Image-Text Retrieval}

We evaluated \model on image-text retrieval tasks on both \dset, and BIMCV-R~\cite{chen2024bimcv}, an external public benchmark of lung CTs and corresponding reports. We computed key retrieval metrics such as recall@k and mean and median rank. For \dset, we computed finding-based metrics, where ``P@5 (F)'' corresponds to how frequently each of the 5 retrieved samples contain at least one common positive finding (F) with the ground-truth match. ``R@5 (F)'' corresponds to the frequency of finding at least one sample containing exactly the same labels as the ground-truth in the top-5 samples.

\begin{figure*}[h]
\centering
  \includegraphics[width=1\linewidth]{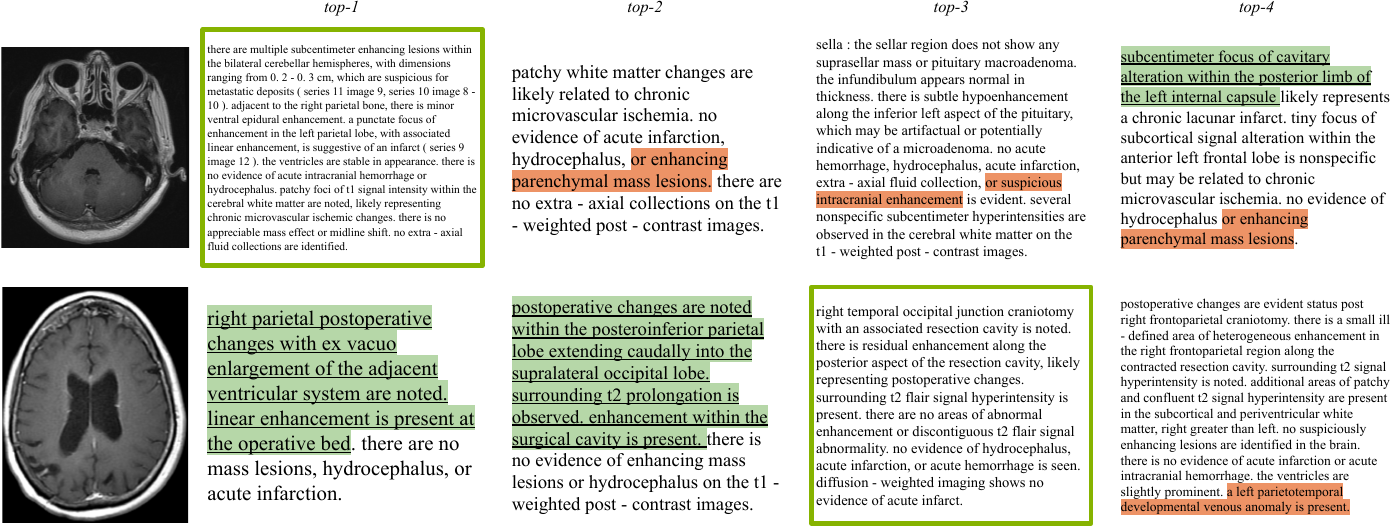}
  \caption{Qualitative examples showing the top-4 output of our \model model for image-to-text retrieval on a reduced dev set of 315 examples. On this subset the median rank achieved was 2. Enboxed examples are correct. Green (and underlined) sections are passages that are clinically correct, even though they are from a different MRI. In red are passages that don't correspond to the MRI.}
  \label{fig:retr_ex}
\end{figure*}

\vspace{-2ex}
\paragraph{\dset} We evaluated \model against multiple baselines: CLIP, Q-Former (the base of \modelS), \model with the traditional Colbert matching algorithm~\cite{khattab2020colbert} instead of PVA, \model without QD, \model with simple pairwise repulsion as used in~\cite{zhang2022multi} instead of DPPs, and a Q-Former with QD. Results with ViT and ResNet-50 backbone models are provided for completeness. For simplicity, only MRIs that have at least one positive finding (around 90\% of our original dataset) were included in the evaluation, as negative reports usually apply to all negative images, making exact matching a faulty metric. As shown in Table~\ref{tab:msk}, except mean rank, \model sets the benchmark on all metrics. The lower mean rank can generally be explained by the model making higher confidence predictions, and this can be adjusted by selecting the weights at earlier training steps or by average pooling the multi-view embeddings at inference instead of using the PVA algorithm. The QD component improved performance, suggesting reliance of PVA on QD for effective learning. Q-Former alone did not show benefits with QD repulsion, suggesting that QD is only effective when query tokens are also encouraged to be aligned with diverse clinical features. We also empirically validate the superiority of DPPs over simple pairwise repulsion. Our approach also outperforms the Colbert algorithm for matching multi-view embeddings. Qualitative examples of images and corresponding reports retrieved by \model are shown in Figure~\ref{fig:retr_ex}.

\vspace{-2ex}
\paragraph{BIMCV-R}

To demonstrate the generalizability of our framework, we also pre-trained it from scratch on the \bim dataset, a publicly available dataset of lung CT scans paired with radiology reports~\cite{chen2024bimcv}. Similar to the original paper, we find that conventional contrastive loss approaches such as a basic Q-Former or CLIP perform very poorly (see Table~\ref{tab:bim}). Notably, without the need for additional SSL techniques used by MedFinder, \model achieves comparable performance purely by leveraging textual supervision. We also identify certain quality issues within the BIMCV-R dataset, detailed in Appendix~\ref{app:bim}, which may contribute to the generally lower performance observed on this benchmark. Despite these limitations, our results show that \model can be effectively applied off-the-shelf to other medical imaging modalities with complex visuals and lengthy reports. We note a greater discrepancy in mean rank on BIMCV-R, but limited methodological details in prior work and unavailable model weights make direct comparison difficult, leaving this to further examination.

\begin{table*}[h]
  \centering
  \resizebox{\textwidth}{!}{%
  \begin{tabular}{l l l c c c c c c c c c c c c}
    \toprule
    Backbone & Pre-training & LLM & \multicolumn{5}{c}{GREEN (LLM Eval)} & \multicolumn{7}{c}{NLG Metrics} \\
    \cmidrule(lr){4-8} \cmidrule(lr){9-15}
             &              &     & All  & FP   & FN   & Location & Severity & METEOR & CIDEr & Rouge-L & Bleu-1 & Bleu-2 & Bleu-3 & Bleu-4 \\
    \midrule
    Densenet-121       & None           & Llama 3.2-1B & 0.300 & 0.110 & 0.190 & 0.750 & 0.850 & 0.117  & 0.039  & 0.180  & 0.177  & 0.103  & 0.065  & 0.042 \\
    Densenet-121 \emph{(fr.)}    & Classification & Llama 3.2-1B & 0.310 & 0.115 & 0.195 & 0.760 & 0.860 & 0.102  & 0.049  & 0.187  & 0.124  & 0.072  & 0.048  & 0.033 \\
    Densenet-121 \emph{(fr.)}    & QFormer        & Llama 3.2-1B & 0.375 & 0.138 & 0.287 & 0.840 & 0.911 & 0.131  & 0.079  & 0.216  & 0.201  & 0.123  & 0.081  & 0.056 \\
    Densenet-121 \emph{(fr.)}    & \model         & Llama 3.2-1B & 0.390 & 0.150 & 0.300 &  \textbf{0.860} &  \textbf{0.920} & \textbf{0.134}  & 0.098  & 0.214  & \textbf{0.241}  & \textbf{0.142}  & \textbf{0.091}  & \textbf{0.061} \\
    QFormer \emph{(fr.)}         & QFormer        & Llama 3.2-1B & 0.360 & 0.130 & 0.280 & 0.820 & 0.900 & 0.125  & 0.105  & 0.210  & 0.190  & 0.115  & 0.078  & 0.053 \\
    QFormer \emph{(fr.)}         & \model         & Llama 3.2-1B & \textbf{0.402} &  \textbf{0.172} &  \textbf{0.318} & 0.852 & 0.917 & 0.128  & \textbf{0.114}  & \textbf{0.219}  & 0.197  & 0.121  & 0.081  & 0.056 \\
    \bottomrule
  \end{tabular}%
}
  \caption{The backbone is always frozen, except for ``None'' pre-training. The GREEN metric is obtained using a 7B parameter LLM. Four GREEN scores, relating to false findings (FP), missing findings (FN), false findings (FP), and accuracy of severity and location specification of findings are provided. Two additional metrics used in GREEN, missing or hallucinated references to prior scans are omitted as we removed these references from our dataset and therefore our models all score a 100\% on these metrics.}
  \label{tab:rgen}
\end{table*}

\subsection{Downstream Tasks}

In this section, we showcase how our pre-training is beneficial for a wide range of downstream tasks.

\vspace{-2ex}
\paragraph{Brain MRI Report Generation.} 

VLP naturally suits radiology report generation, as the visual embeddings already align with text features. We evaluated our pre-trained backbones by freezing the vision backbone and assessed how well a language model extracted image-grounded information. We used Llama-3.2-1B~\cite{grattafiori2024llama}, providing either multi-view embeddings or image feature maps to the LLM via a bridging MLP. Training and evaluation were done on \dset. We computed both LLM-based metrics (GREEN metric~\cite{ostmeier2024green}) and natural language generation metrics. We compared \model to training from scratch, Q-Former pre-training, and classification-based pre-training (``CLS''), using either the vision backbone or multi-view embeddings as LLM input. CLS training was done via the labels we extracted from our reports using the methodology described in Appendix~\ref{app:gpt4}. As shown in Table~\ref{tab:rgen}, VLP improved over no pre-training or classification pre-training. \model also leads to additional improvements over simple QFormer pre-training. This is the first work to provide report generation capabilities for brain MRIs that were trained on a large-scale dataset. Our results demonstrate that the ability of VLMs to generate reports for brain MRIs is in line with other radiographic modalities, such as chest X-rays. Example reports are shown in Figure~\ref{fig:gen-example}.

\vspace{-2ex}
\paragraph{Alzheimer Classification.}

\begin{figure}[h]
  \centering
   \includegraphics[width=1\linewidth]{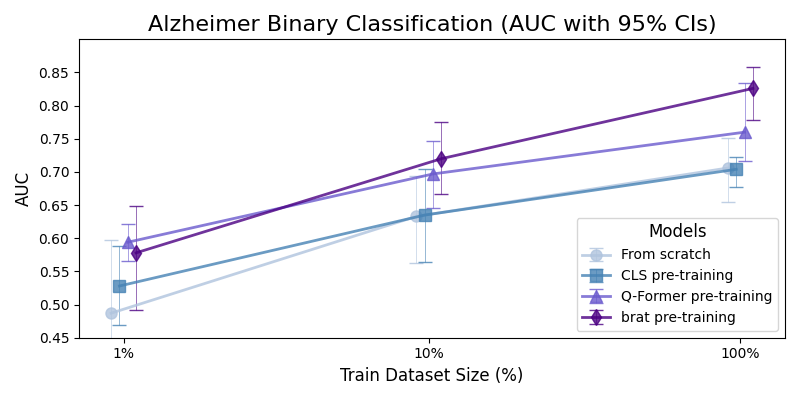}
   \caption{Comparison of \model pre-training to alternative pre-training methods for Alzheimer's disease classification on ADNI.}
   \label{fig:adni}
\end{figure}

To investigate whether our pre-training generalizes to non-cancer-focused brain MRI datasets, we evaluated \model on ADNI~\cite{petersen2010alzheimer}, a dataset for investigating the progression of Alzheimer's disease. We split the cohort into training (n=1,932), validation (n=384), and hold-out test (n=291) sets. Brain MRIs are either ``cognitive normal'', ``mild cognitive impairment'' (MCI), or ``Alzheimer's disease''. Figure~\ref{fig:adni} shows the performance on binary classification (Normal or Alz.) for 1, 10, and 100\% training data. As results on ADNI vary significantly based on random seeds and selected subsets of the training data, we launched 10 runs for each setting and bootstrap from these results to obtain 95\% confidence intervals. Table~\ref{tab:adni_results} in the Appendix contains more extensive results, which shows that our pre-training approach led to consistently accurate performance across all analyzed settings. VLP, in general, produced clear performance improvements over classification pre-training. These results underline the usefulness of \model across non-cancer domains.

\vspace{-2ex}
\paragraph{Segmentation Tasks.}

\begin{figure}[h]
  \centering
   \includegraphics[width=1\linewidth]{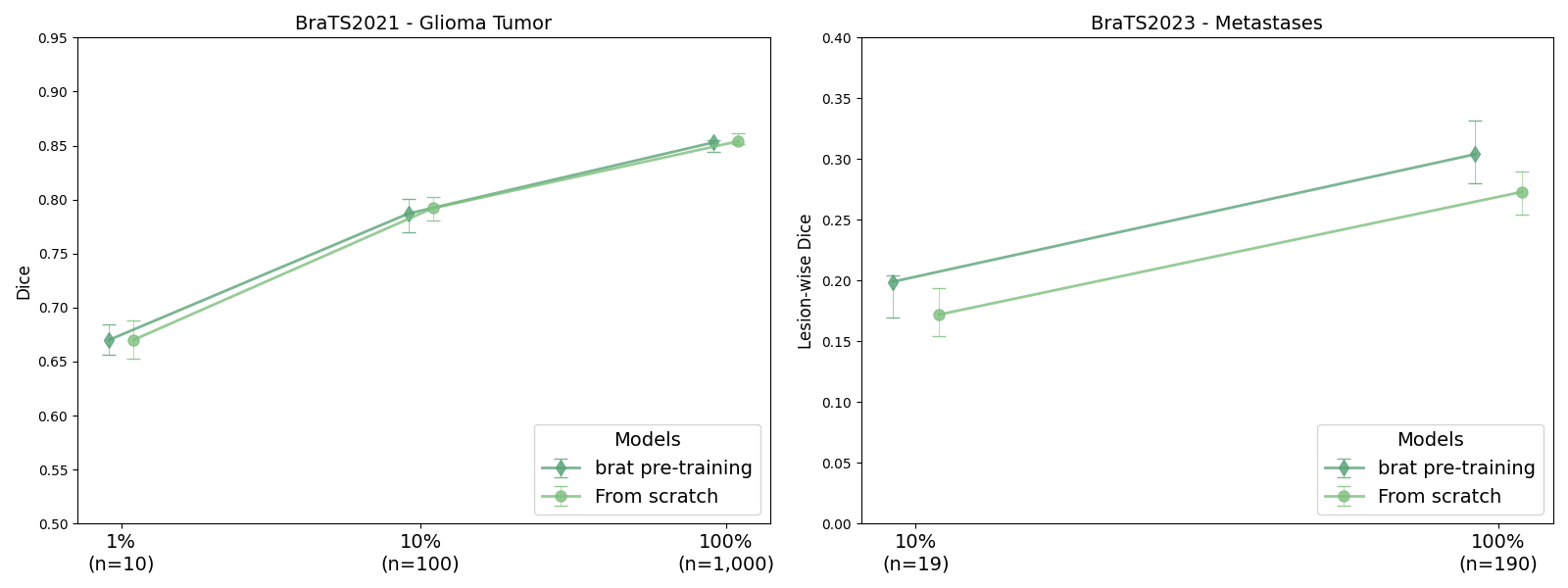}
   \caption{Comparison of \model pre-training to scratch training for tumor (left) and metastases (right) segmentation on BraTS2021 and BraTS2023. Scores are averaged across the three tumor regions.}
   \label{fig:brats}
\end{figure}

We also evaluated \model on one of the most common downstream applications in brain MRI analysis: tumor segmentation. We used BraTS2021~\cite{baid2021rsna}, containing gliomas, and BraTS2023-METS~\cite{moawad2023brain}, containing brain metastases. In order to isolate the benefit of the \model pre-training framework, we only used T1W MRI and did not include some of the post-processing steps typically included for these datasets. We employed 4-fold cross-validation following conventional methods~\cite{isensee2024nnu}. In addition, three random seeds were used for each run to obtain confidence intervals. Benchmark relevant evaluation metrics including the Dice (Brats2021) and lesion-wise Dice (Brats2023) were computed from three overlapping regions, namely whole tumor, tumor core, and enhancing tumor. Figure~\ref{fig:brats} shows that our pre-training improves performance for the metastases, but not for the gliomas. Of note, tumor core is easily separated from its background due to higher soft-tissue contrast, and can be identifed even by non-expert obviating the need for precise anatomical understanding of brain MRIs. More detailed results are provided in Appendix Table~\ref{tab:brats2021_results} and~\ref{tab:brats2023_results}.

\section{Conclusion}

We present a large-scale dataset and introduce two ideas novel to vision-language representation learning: multi-view embeddings, enabled via the PVA algorithm, and DPPs to maximize the quality-diversity of these embeddings. Our approach demonstrates promising results when applied to images paired with long reports, including both brain MRI and lung CT datasets. The proposed \model framework is architecture-agnostic and compatible with a variety of image and text encoders. The flexibility of the learnable multi-view embeddings also naturally allows to extend the input beyond imaging data. This is promising for medical imaging, where patient context and lab results influence diagnosis.
{
    \small
    \bibliographystyle{ieeenat_fullname}
    \bibliography{main}

@String(ICLR = {Int. Conf. Learn. Represent.})

@String(AAAI = {AAAI})

@String(ICLR  = {ICLR})

@inproceedings{radford2021learning,
  title={Learning transferable visual models from natural language supervision},
  author={Radford, Alec and Kim, Jong Wook and Hallacy, Chris and Ramesh, Aditya and Goh, Gabriel and Agarwal, Sandhini and Sastry, Girish and Askell, Amanda and Mishkin, Pamela and Clark, Jack and others},
  booktitle={International Conference on Machine Learning},
  year={2021},
  organization={PMLR}
}

@inproceedings{chen2020simple,
  title={A simple framework for contrastive learning of visual representations},
  author={Chen, Ting and Kornblith, Simon and Norouzi, Mohammad and Hinton, Geoffrey},
  booktitle={International Conference on Machine Learning},
  year={2020}
}

@inproceedings{beit2022,
title={{BEiT}: {BERT} Pre-Training of Image Transformers},
author={Hangbo Bao and Li Dong and Songhao Piao and Furu Wei},
booktitle={International Conference on Learning Representations},
year={2022}
}

@inproceedings{he2022masked,
  title={Masked autoencoders are scalable vision learners},
  author={He, Kaiming and Chen, Xinlei and Xie, Saining and Li, Yanghao and Doll{\'a}r, Piotr and Girshick, Ross},
  booktitle={Proceedings of the IEEE/CVF conference on computer vision and pattern recognition},
  pages={16000--16009},
  year={2022}
}

@inproceedings{tang2022self,
  title={Self-supervised pre-training of swin transformers for 3d medical image analysis},
  author={Tang, Yucheng and Yang, Dong and Li, Wenqi and Roth, Holger R and Landman, Bennett and Xu, Daguang and Nath, Vishwesh and Hatamizadeh, Ali},
  booktitle={Proceedings of the IEEE/CVF conference on computer vision and pattern recognition},
  pages={20730--20740},
  year={2022}
}

@inproceedings{jiang2022self,
  title={Self-supervised 3D anatomy segmentation using self-distilled masked image transformer (SMIT)},
  author={Jiang, Jue and Tyagi, Neelam and Tringale, Kathryn and Crane, Christopher and Veeraraghavan, Harini},
  booktitle={International Conference on Medical Image Computing and Computer-Assisted Intervention},
  pages={556--566},
  year={2022},
  organization={Springer}
}

@inproceedings{vincent2008extracting,
  title={Extracting and composing robust features with denoising autoencoders},
  author={Vincent, Pascal and Larochelle, Hugo and Bengio, Yoshua and Manzagol, Pierre-Antoine},
  booktitle={International Conference on Machine Learning},
  year={2008}
}

@inproceedings{zhang2016colorful,
  title={Colorful image colorization},
  author={Zhang, Richard and Isola, Phillip and Efros, Alexei A},
  booktitle={European Conference on Computer Vision},
  year={2016},
  organization={Springer}
}

@inproceedings{noroozi2016unsupervised,
  title={Unsupervised learning of visual representations by solving jigsaw puzzles},
  author={Noroozi, Mehdi and Favaro, Paolo},
  booktitle={European Conference on Computer Vision},
  year={2016},
  organization={Springer}
}

@article{munk2024amaes,
  title={AMAES: Augmented Masked Autoencoder Pretraining on Public Brain MRI Data for 3D-Native Segmentation},
  author={Munk, Asbjørn and Ambsdorf, Jakob and Llambias, Sebastian and Nielsen, Mads},
  journal={arXiv preprint arXiv:2408.00640},
  year={2024}
}

@inproceedings{ostmeier2024green,
  title={GREEN: Generative Radiology Report Evaluation and Error Notation},
  author={Ostmeier, Sophie and Xu, Justin and Chen, Zhihong and Varma, Maya and Blankemeier, Louis and Bluethgen, Christian and Md, Arne and Moseley, Michael and Langlotz, Curtis and Chaudhari, Akshay and others},
  booktitle={Findings of the Association for Computational Linguistics: EMNLP 2024},
  pages={374--390},
  year={2024}
}

@article{grattafiori2024llama,
  title={The llama 3 herd of models},
  author={Grattafiori, Aaron and Dubey, Abhimanyu and Jauhri, Abhinav and Pandey, Abhinav and Kadian, Abhishek and Al-Dahle, Ahmad and Letman, Aiesha and Mathur, Akhil and Schelten, Alan and Vaughan, Alex and others},
  journal={arXiv preprint arXiv:2407.21783},
  year={2024}
}

@InProceedings{Huang_2017_CVPR,
author = {Huang, Gao and Liu, Zhuang and van der Maaten, Laurens and Weinberger, Kilian Q.},
title = {Densely Connected Convolutional Networks},
booktitle = {Proceedings of the IEEE/CVF Conference on Computer Vision and Pattern Recognition},
year = {2017}
}

@inproceedings{zhang2022multi,
  title={Multi-View Document Representation Learning for Open-Domain Dense Retrieval},
  author={Zhang, Shunyu and Liang, Yaobo and Gong, Ming and Jiang, Daxin and Duan, Nan},
  booktitle={Proceedings of the 60th Annual Meeting of the Association for Computational Linguistics (Volume 1: Long Papers)},
  year={2022}
}

@inproceedings{kovaleva2019revealing,
  title={Revealing the Dark Secrets of BERT},
  author={Kovaleva, Olga and Romanov, Alexey and Rogers, Anna and Rumshisky, Anna},
  booktitle={Proceedings of the 2019 Conference on Empirical Methods in Natural Language Processing and the 9th International Joint Conference on Natural Language Processing (EMNLP-IJCNLP)},
  year={2019}
}

@inproceedings{clark-etal-2019-bert,
    title = "What Does {BERT} Look at? An Analysis of {BERT}{'}s Attention",
    author = "Clark, Kevin  and
      Khandelwal, Urvashi  and
      Levy, Omer  and
      Manning, Christopher D.",
    booktitle = "Proceedings of the 2019 ACL Workshop BlackboxNLP: Analyzing and Interpreting Neural Networks for NLP",
    year = "2019",
}

@inproceedings{li2023blip,
  title={Blip-2: Bootstrapping language-image pre-training with frozen image encoders and large language models},
  author={Li, Junnan and Li, Dongxu and Savarese, Silvio and Hoi, Steven},
  booktitle={International Conference on Machine Learning},
  year={2023},
}

@article{kulesza2012determinantal,
  title={Determinantal point processes for machine learning},
  author={Kulesza, Alex and Taskar, Ben and others},
  journal={Foundations and Trends{\textregistered} in Machine Learning},
  year={2012}
}

@inproceedings{karpukhin2020dense,
  title={Dense Passage Retrieval for Open-Domain Question Answering},
  author={Karpukhin, Vladimir and Oguz, Barlas and Min, Sewon and Lewis, Patrick and Wu, Ledell and Edunov, Sergey and Chen, Danqi and Yih, Wen-tau},
  booktitle={Proceedings of the 2020 Conference on Empirical Methods in Natural Language Processing (EMNLP)},
  year={2020}
}

@inproceedings{ni2022large,
  title={Large Dual Encoders Are Generalizable Retrievers},
  author={Ni, Jianmo and Qu, Chen and Lu, Jing and Dai, Zhuyun and Abrego, Gustavo Hernandez and Ma, Ji and Zhao, Vincent and Luan, Yi and Hall, Keith and Chang, Ming-Wei and others},
  booktitle={Proceedings of the 2022 Conference on Empirical Methods in Natural Language Processing},
  year={2022}
}

@inproceedings{khattab2020colbert,
  title={Colbert: Efficient and effective passage search via contextualized late interaction over bert},
  author={Khattab, Omar and Zaharia, Matei},
  booktitle={Proceedings of the 43rd International ACM SIGIR conference on research and development in Information Retrieval},
  year={2020}
}

@inproceedings{santhanam2022colbertv2,
  title={ColBERTv2: Effective and Efficient Retrieval via Lightweight Late Interaction},
  author={Santhanam, Keshav and Khattab, Omar and Saad-Falcon, Jon and Potts, Christopher and Zaharia, Matei},
  booktitle={Proceedings of the 2022 Conference of the North American Chapter of the Association for Computational Linguistics: Human Language Technologies},
  year={2022}
}

@inproceedings{cherti2023reproducible,
  title={Reproducible scaling laws for contrastive language-image learning},
  author={Cherti, Mehdi and Beaumont, Romain and Wightman, Ross and Wortsman, Mitchell and Ilharco, Gabriel and Gordon, Cade and Schuhmann, Christoph and Schmidt, Ludwig and Jitsev, Jenia},
  booktitle={Proceedings of the IEEE/CVF Conference on Computer Vision and Pattern Recognition},
  year={2023}
}

@article{sun2023eva,
  title={Eva-clip: Improved training techniques for clip at scale},
  author={Sun, Quan and Fang, Yuxin and Wu, Ledell and Wang, Xinlong and Cao, Yue},
  journal={arXiv preprint arXiv:2303.15389},
  year={2023}
}

@inproceedings{alayrac2022flamingo,
  title={Flamingo: a visual language model for few-shot learning},
  author={Alayrac, Jean-Baptiste and Donahue, Jeff and Luc, Pauline and Miech, Antoine and Barr, Iain and Hasson, Yana and Lenc, Karel and Mensch, Arthur and Millican, Katherine and Reynolds, Malcolm and others},
  booktitle={Advances in Neural Information Processing Systems},
  year={2022}
}

@inproceedings{chenpali,
  title={PaLI: A Jointly-Scaled Multilingual Language-Image Model},
  author={Chen, Xi and Wang, Xiao and Changpinyo, Soravit and Piergiovanni, AJ and Padlewski, Piotr and Salz, Daniel and Goodman, Sebastian and Grycner, Adam and Mustafa, Basil and Beyer, Lucas and others},
  booktitle={International Conference on Learning Representations},
  year={2023}
}

@article{wanggit,
  title={GIT: A Generative Image-to-text Transformer for Vision and Language},
  author={Wang, Jianfeng and Yang, Zhengyuan and Hu, Xiaowei and Li, Linjie and Lin, Kevin and Gan, Zhe and Liu, Zicheng and Liu, Ce and Wang, Lijuan},
  journal={Transactions on Machine Learning Research},
  year={2022}
}

@inproceedings{jaegle2021perceiver,
  title={Perceiver: General perception with iterative attention},
  author={Jaegle, Andrew and Gimeno, Felix and Brock, Andy and Vinyals, Oriol and Zisserman, Andrew and Carreira, Joao},
  booktitle={International Conference on Machine Learning},
  year={2021},
}

@article{johnson2019mimic,
  title={MIMIC-CXR, a de-identified publicly available database of chest radiographs with free-text reports},
  author={Johnson, Alistair EW and Pollard, Tom J and Berkowitz, Seth J and Greenbaum, Nathaniel R and Lungren, Matthew P and Deng, Chih-ying and Mark, Roger G and Horng, Steven},
  journal={Scientific data},
  year={2019}
}

@article{bustos_padchest_2020,
  title = {{{PadChest}}: {{A}} Large Chest x-Ray Image Dataset with Multi-Label Annotated Reports},
  shorttitle = {{{PadChest}}},
  author = {Bustos, Aurelia and Pertusa, Antonio and Salinas, Jose-Maria and de la Iglesia-Vayá, Maria},
  year = {2020},
  journal = {Medical Image Analysis},
}

@inproceedings{irvin_chexpert_2019,
  title={Chexpert: A large chest radiograph dataset with uncertainty labels and expert comparison},
  author={Irvin, Jeremy and Rajpurkar, Pranav and Ko, Michael and Yu, Yifan and Ciurea-Ilcus, Silviana and Chute, Chris and Marklund, Henrik and Haghgoo, Behzad and Ball, Robyn and Shpanskaya, Katie and others},
  booktitle={Proceedings of the AAAI conference on artificial intelligence},
  year={2019}
}

@article{chauhan2024continuous,
  title={Continuous patient state attention model for addressing irregularity in electronic health records},
  author={Chauhan, Vinod Kumar and Thakur, Anshul and O’Donoghue, Odhran and Rohanian, Omid and Molaei, Soheila and Clifton, David A},
  journal={BMC Medical Informatics and Decision Making},
  year={2024},
}

@inproceedings{sharma2018conceptual,
  title={Conceptual captions: A cleaned, hypernymed, image alt-text dataset for automatic image captioning},
  author={Sharma, Piyush and Ding, Nan and Goodman, Sebastian and Soricut, Radu},
  booktitle={Proceedings of the 56th Annual Meeting of the Association for Computational Linguistics},
  year={2018}
}

@inproceedings{liu2023exploring,
  title={Exploring the Boundaries of GPT-4 in Radiology},
  author={Liu, Qianchu and Hyland, Stephanie and Bannur, Shruthi and Bouzid, Kenza and Castro, Daniel and Wetscherek, Maria and Tinn, Robert and Sharma, Harshita and P{\'e}rez-Garc{\'\i}a, Fernando and Schwaighofer, Anton and others},
  booktitle={Proceedings of the 2023 Conference on Empirical Methods in Natural Language Processing},
  year={2023}
}

@inproceedings{zhang2022contrastive,
  title={Contrastive learning of medical visual representations from paired images and text},
  author={Zhang, Yuhao and Jiang, Hang and Miura, Yasuhide and Manning, Christopher D and Langlotz, Curtis P},
  booktitle={Machine Learning for Healthcare Conference},
  year={2022}}

@inproceedings{huang_gloria_2021,
  title={Gloria: A multimodal global-local representation learning framework for label-efficient medical image recognition},
  author={Huang, Shih-Cheng and Shen, Liyue and Lungren, Matthew P and Yeung, Serena},
  booktitle={Proceedings of the IEEE/CVF International Conference on Computer Vision},
  year={2021}
}

@inproceedings{bardes2022vicreg,
  title={VICREG: VARIANCE-INVARIANCE-COVARIANCE REGULARIZATION FOR SELF-SUPERVISED LEARNING},
  author={Bardes, Adrien and Ponce, Jean and LeCun, Yann},
  booktitle={10th International Conference on Learning Representations, ICLR 2022},
  year={2022}
}

@inproceedings{saran2023streaming,
  title={Streaming active learning with deep neural networks},
  author={Saran, Akanksha and Yousefi, Safoora and Krishnamurthy, Akshay and Langford, John and Ash, Jordan T},
  booktitle={International Conference on Machine Learning},
  year={2023},
}

@article{baid2021rsna,
  title={The rsna-asnr-miccai brats 2021 benchmark on brain tumor segmentation and radiogenomic classification},
  author={Baid, Ujjwal and Ghodasara, Satyam and Mohan, Suyash and Bilello, Michel and Calabrese, Evan and Colak, Errol and Farahani, Keyvan and Kalpathy-Cramer, Jayashree and Kitamura, Felipe C and Pati, Sarthak and others},
  journal={arXiv preprint arXiv:2107.02314},
  year={2021}
}

@article{petersen2010alzheimer,
  title={Alzheimer's disease Neuroimaging Initiative (ADNI) clinical characterization},
  author={Petersen, Ronald Carl and Aisen, Paul S and Beckett, Laurel A and Donohue, Michael C and Gamst, Anthony Collins and Harvey, Danielle J and Jack Jr, CR and Jagust, William J and Shaw, Leslie M and Toga, Arthur W and others},
  journal={Neurology},
  year={2010},
}

@inproceedings{isensee2024nnu,
  title={nnu-net revisited: A call for rigorous validation in 3d medical image segmentation},
  author={Isensee, Fabian and Wald, Tassilo and Ulrich, Constantin and Baumgartner, Michael and Roy, Saikat and Maier-Hein, Klaus and Jaeger, Paul F},
  booktitle={International Conference on Medical Image Computing and Computer-Assisted Intervention},
  year={2024},
}

@article{moawad2023brain,
  title={The brain tumor segmentation (brats-mets) challenge 2023: Brain metastasis segmentation on pre-treatment mri},
  author={Moawad, Ahmed W and Janas, Anastasia and Baid, Ujjwal and Ramakrishnan, Divya and Saluja, Rachit and Ashraf, Nader and Jekel, Leon and Amiruddin, Raisa and Adewole, Maruf and Albrecht, Jake and others},
  journal={arXiv preprint arXiv:2306.00838},
  year={2023}
}

@article{li2021exploring,
  title={Exploring global diverse attention via pairwise temporal relation for video summarization},
  author={Li, Ping and Ye, Qinghao and Zhang, Luming and Yuan, Li and Xu, Xianghua and Shao, Ling},
  journal={Pattern Recognition},
  year={2021},
  publisher={Elsevier}
}

@inproceedings{zbontar2021barlow,
  title={Barlow twins: Self-supervised learning via redundancy reduction},
  author={Zbontar, Jure and Jing, Li and Misra, Ishan and LeCun, Yann and Deny, St{\'e}phane},
  booktitle={International conference on machine learning},
  year={2021},
}

@inproceedings{wang2022multi,
  title={Multi-granularity cross-modal alignment for generalized medical visual representation learning},
  author={Wang, Fuying and Zhou, Yuyin and Wang, Shujun and Vardhanabhuti, Varut and Yu, Lequan},
  booktitle={Advances in Neural Information Processing Systems},
  year={2022}
}

@inproceedings{wu2023medklip,
  title={Medklip: Medical knowledge enhanced language-image pre-training for x-ray diagnosis},
  author={Wu, Chaoyi and Zhang, Xiaoman and Zhang, Ya and Wang, Yanfeng and Xie, Weidi},
  booktitle={Proceedings of the IEEE/CVF International Conference on Computer Vision},
  year={2023}
}

@article{zhang2023knowledge,
  title={Knowledge-enhanced visual-language pre-training on chest radiology images},
  author={Zhang, Xiaoman and Wu, Chaoyi and Zhang, Ya and Xie, Weidi and Wang, Yanfeng},
  journal={Nature Communications},
  year={2023},
  publisher={Nature Publishing Group UK London}
}

@article{yang2024advancing,
  title={Advancing multimodal medical capabilities of Gemini},
  author={Yang, Lin and Xu, Shawn and Sellergren, Andrew and Kohlberger, Timo and Zhou, Yuchen and Ktena, Ira and Kiraly, Atilla and Ahmed, Faruk and Hormozdiari, Farhad and Jaroensri, Tiam and others},
  journal={arXiv preprint arXiv:2405.03162},
  year={2024}
}

@inproceedings{chen2024bimcv,
  title={Bimcv-r: A landmark dataset for 3d ct text-image retrieval},
  author={Chen, Yinda and Liu, Che and Liu, Xiaoyu and Arcucci, Rossella and Xiong, Zhiwei},
  booktitle={International Conference on Medical Image Computing and Computer-Assisted Intervention},
  year={2024},
}

@inproceedings{windsorvision,
  title={Vision-Language Modelling For Radiological Imaging and Reports In The Low Data Regime},
  author={Windsor, Rhydian and Jamaludin, Amir and Kadir, Timor and Zisserman, Andrew},
  booktitle={Medical Imaging with Deep Learning},
  year={2023}
}

@inproceedings{konwer2023enhancing,
  title={Enhancing modality-agnostic representations via meta-learning for brain tumor segmentation},
  author={Konwer, Aishik and Hu, Xiaoling and Bae, Joseph and Xu, Xuan and Chen, Chao and Prasanna, Prateek},
  booktitle={Proceedings of the IEEE/CVF International Conference on Computer Vision},
  year={2023}
}

@inproceedings{xufeasibility,
  title={Feasibility and benefits of joint learning from MRI databases with different brain diseases and modalities for segmentation},
  author={Xu, Wentian and Moffat, Matthew and Seale, Thalia and Liang, Ziyun and Wagner, Felix and Whitehouse, Daniel and Menon, David and Newcombe, Virginia and Voets, Natalie and Banerjee, Abhirup and others},
  booktitle={Medical Imaging with Deep Learning},
  year={2024}
}
}
\clearpage
\setcounter{page}{1}
\maketitlesupplementary


\section{Dataset Details} \label{app:dset}

In this section, we discuss \dset, the largest existing dataset of brain MRIs and radiology reports.

\subsection{Raw Dataset}

We collected the brain MRI scans and their corresponding clinical reports from a cancer center, covering the period from 2012 to 2017. These scans were primarily obtained to monitor brain metastases and tumours in cancer patients, resulting in a dataset rich in positive findings (89.7\% of scans show abnormalities, an average of 134 words or 8 sentences per report) and representative of a diverse patient population. We also collected the clinical reports corresponding to these images, as well as extensive demographic data, primary diagnosis, ongoing chemotherapy and radiotherapy treatments, and survival information, which will be utilized in future studies. 

\begin{figure*}[t!]
    \centering
    \begin{tabular}{cc}
        \includegraphics[width=0.49\textwidth]{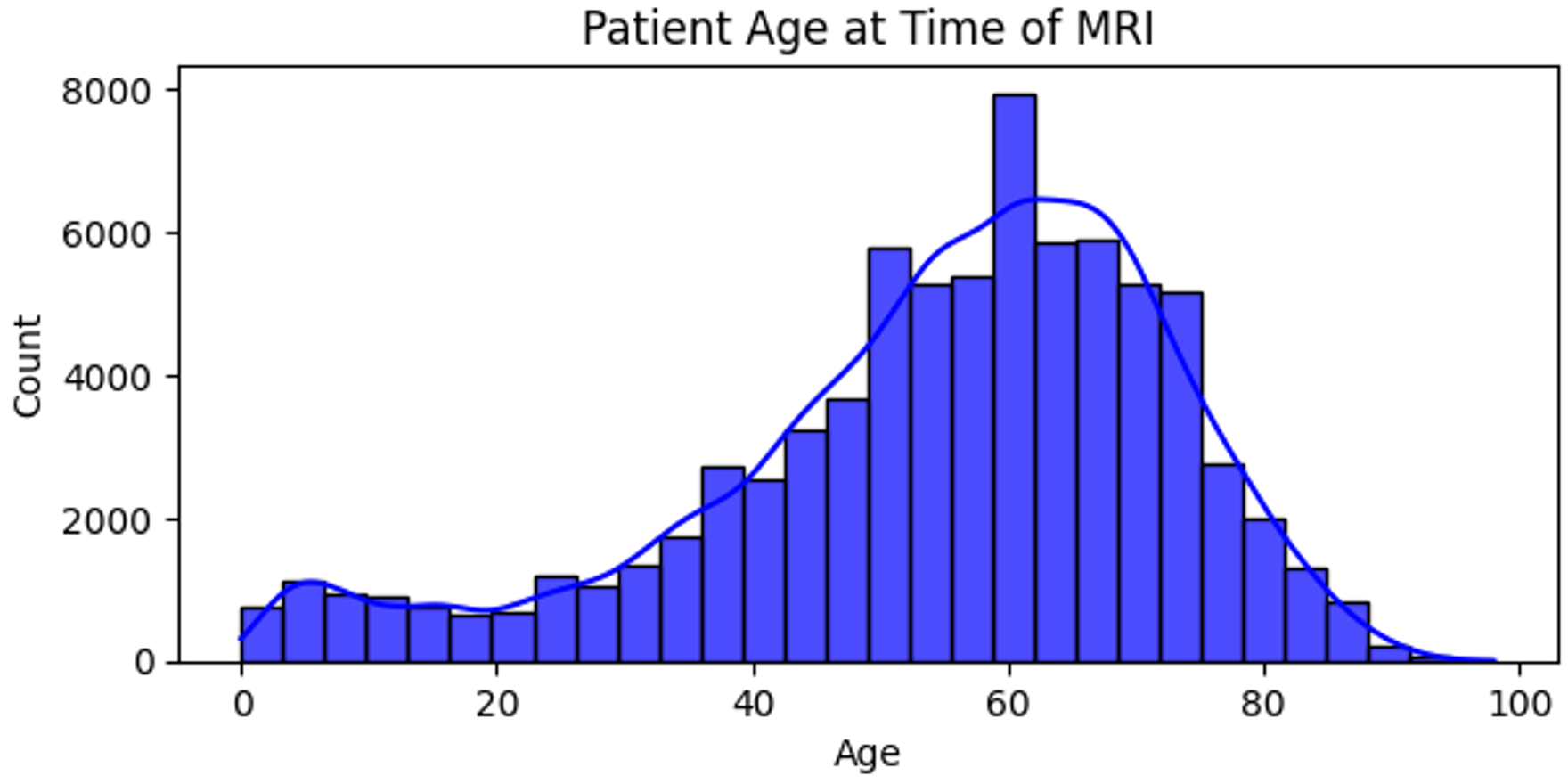} &
        \includegraphics[width=0.49\textwidth]{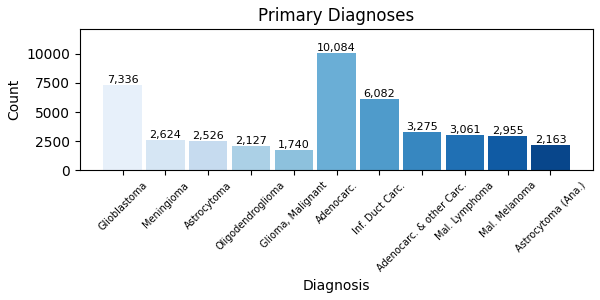} \\
        \includegraphics[width=0.49\textwidth]{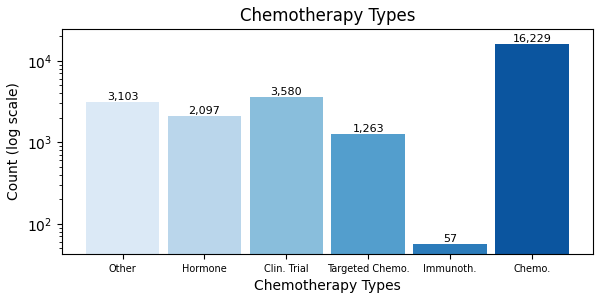} &
        \includegraphics[width=0.49\textwidth]{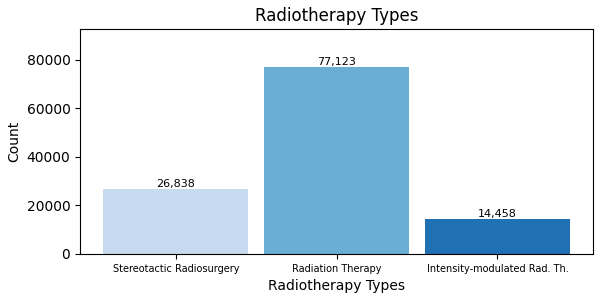} 
    \end{tabular}
    \caption{Participant demographics. Primary diagnoses refers to the primary cancer diagnosis for the patients for whomst the scan was ordered. Chemotherapy and radiotherapy types show a count of all the types of chemo/radio sessions assigned to the patients in this dataset.}
    \label{fig:brain-demo}
\end{figure*}

MRI sessions typically consist of multiple MRI modality scans (e.g. FLAIR, T2, etc.); however, in this first iteration, we focus on T1-post contrast MRIs, the most informative type of scan. To extract T1 post-contrast scans, we generated a long, clinician-validated list of keywords typically used to refer to these scans. The list contained over 50 expressions such as ``Axial T1 post SENSE'' or ``Ax T1 POST''. We removed around 3,000 sessions that did not include T1 post-contrast imaging. DICOM medical images were converted and saved as 3D NIFTI files. We first sorted and loaded the DICOM slices into a 3D array, and then, for each image stack, preserved relevant metadata, such as pixel spacing and slice thickness, in the affine matrix. To standardize the intensity values, we thresholded the images at the 99th percentile and rescaled them to a range of 0-800, converting the final values to 16-bit integers. We saved the processed 3D volumes as compressed NIFTI files (.nii.gz). We then connect the MRI sessions to patient data stored in a REDCap database. For each imaging session, we attached the patient's demographic information and added calculated fields like age and time to death. We then mapped treatments and diagnoses to imaging sessions by finding the closest diagnosis date and checking which medications and radiation therapies were active at the time of imaging. We provide an overview of key patient data in Figure~\ref{fig:brain-demo}. The final dataset includes 77,228 brain MRI image-report pairs from 24,262 unique patients. To develop our model, we performed a patient-wise split of the data into 75,142 examples for training, 945 for development, and 1,141 for the test set.

\subsection{Text Processing} \label{app:gpt4}

\begin{figure}[h!]
    \small
    \begin{tcolorbox}[
        colback=gray!5,    
        boxrule=0.5pt,     
        arc=0pt            
    ]
    {\colorbox{cyan!20}{Again}} status post left-sided craniotomy with stable postoperative changes and with slight {\colorbox{cyan!20}{increase}} in the heterogeneously enhancing mass lesion centred in the left temporal lobe which {\colorbox{cyan!20}{now measures}} 7.5 x 4.8 cm on image 13 series 14 {\colorbox{cyan!20}{from 6.7 x 4.7 cm}}, though the enhancement within it is more irregular and less intense {\colorbox{cyan!20}{than before}}. {\colorbox{yellow!30}{The mass is not completely imaged on the perfusion seq-}} {\colorbox{yellow!30}{uence but there is hyperperfusion inferiorly within the }} {\colorbox{yellow!30}{nodular enhancing component which is incompletely de-}} {\colorbox{yellow!30}{monstrated}}. The surrounding hyperintense T2/FLAIR infiltrating nonenhancing signal abnormality is stable consistent with nonenhancing tumor/edema. {\colorbox{cyan!20}{No new}} discontinuous suspiciously enhancing brain lesions. There is slightly {\colorbox{cyan!20}{increased}} dilatation of the ventricles with slightly {\colorbox{cyan!20}{increased}} hyperintense T2/FLAIR signal in the periventricular white matter particularly about the frontal horns and atrium, suggesting transependymal flow of CSF from a communicating hydrocephalus. Stable mild midline shift to the right without significant downward herniation. No acute intracranial hemorrhage, infarct, or new extra-axial collections.
    \end{tcolorbox}
    \caption[Raw report example]{An example report showing references to prior scans in {\colorbox{cyan!20}{blue}} and descriptions of findings not visible on T1 post-contrast scans in {\colorbox{yellow!30}{yellow}}.}
    \label{fig:brain-gpt-moti}
\end{figure}

\begin{figure}[h]
    \centering
    \includegraphics[width=0.5\linewidth]{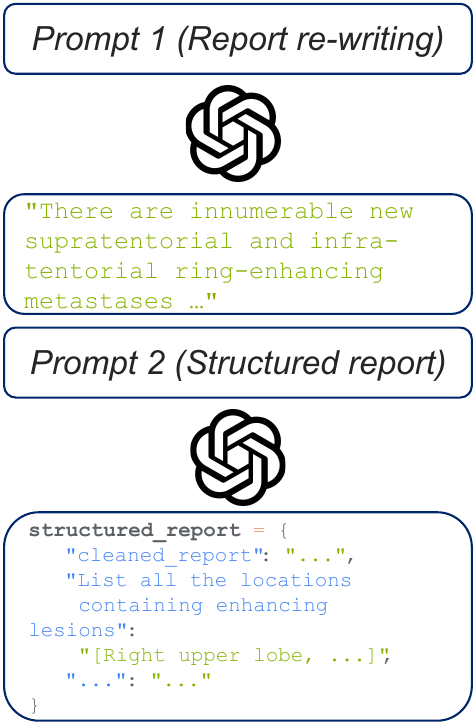}
    \caption{The two-step GPT-4 based report processing pipeline. 
    Prompt 1 and 2 are in Figure~\ref{fig:brain-prompt1} and~\ref{fig:brain-prompt2}, respectively.}
    \label{fig:brain-gpt-proc}
\end{figure}

As this paper focuses on learning image representations from brain MRIs, we had to ensure that all report content was visually grounded in the corresponding images. For example, keyword filtering revealed that 94\% of reports make references to prior scans. Figure~\ref{fig:brain-gpt-moti} contains an example report showing the two main types of information not detectable from the image: references to prior scans and description of findings not visible on T1 post-contrast images. 

\begin{figure}[h]
    \small
    \begin{tcolorbox}[
        colback=gray!5,    
        boxrule=0.5pt,     
        arc=0pt            
    ]
    You are a highly experienced radiologist. Re-write the given brain MRI report and only modify the following: \\ \\
    (a) Leave out any details not visible on T1-weighted post-contrast images. Note that T2/FLAIR hyperintensities can often be seen on T1 Images. Observations related to e.g. perfusion, plasma volume or K trans cannot be seen and should be excluded. \\
    (b) Leave out any terms that suggest temporal change or progression (e.g. dates, ``new'', ``increased'', ``previous'', ``now'', ``compared to'', ``since last'', ``more'', ``less'', etc.) \\
    (c) Remove any PHI.
    \end{tcolorbox}
    \caption[Report re-writing prompt]{The final prompt that was used to re-write the reports and remove PHI and information not visible from the T1 post-contrast images.}
    \label{fig:brain-prompt1}
\end{figure}

Inspired by recent work demonstrating GPT-4 performs well on radiology report processing~\citep{liu2023exploring}, we developed an on-premise GPT-4-based report processing pipeline. This pipeline enabled us to achieve three objectives without using expensive human annotators: the data was anonymized by removing protected health information (PHI), the reports were re-written to remove the aforementioned references, and we extracted structured information. Through iterative prompt engineering and radiologist feedback, we arrived at a 2-stage approach that extracted information from the reports with a 96\% accuracy on a gold standard set of 50 manually annotated reports. This accuracy was calculated by counting the share of exact matches for all structured data points. By evaluating the second step of the two-step approach, we also implicitly ensure that the intermediate re-written report captures all the relevant information from the original report. 

\begin{figure*}[h!]
\centering
\begin{tikzpicture}[remember picture, scale=0.5, transform shape]

\node[anchor=north, text width=2\textwidth] (box1) at (0, 0) {
    \begin{tcolorbox}[title=Original Report, width=\textwidth, colframe=blue!50!black, colback=blue!5, fonttitle=\bfseries]
    FINDINGS: BRAIN: The patient is status post prior left anterior parietal craniotomy. Slightly increased heterogeneously enhancing nodularity along the posterior medial aspect of the left frontal operative bed as seen on series 4 image 26 measuring 1.5 x 1.1 cm, previously 1.4 x 1.1 cm on July 5, 2016 and 1.3 x 0.9 cm on January 4, 2016. No ependymal or leptomeningeal foci have developed. Perfusion series again demonstrates elevated kTrans and plasma volume within this nodular focus. No hydrocephalus, midline shift or herniation. No at parenchymal hemorrhage. No abnormal extra-axial collections. No acute infarct. Normal intracranial arterial flow-voids. \\
    IMPRESSION: Subtle increase in size of the nodular enhancing tumor along the posterior medial aspect of the left middle frontal operative bed, the tumor demonstrating focal elevation of perfusion parameters, a characteristic finding with oligodendrogliomas.
    \end{tcolorbox}
};

\node[anchor=north] (arrow1) at ([yshift=0.1cm]box1.south) {
    \huge $\downarrow$
};

\node[anchor=north, text width=2\textwidth] (box2) at ([yshift=-0.4cm]box1.south) {
    \begin{tcolorbox}[title=Rewritten Report, width=\textwidth, colframe=blue!40!black, colback=blue!10, fonttitle=\bfseries]
    The patient has undergone a left anterior parietal craniotomy. There is a heterogeneously enhancing nodularity along the posterior medial aspect of the left frontal operative bed, visible on series 4 image 26, measuring 1.5 x 1.1 cm. No ependymal or leptomeningeal foci are present. No hydrocephalus, midline shift or herniation is observed. There is no parenchymal hemorrhage. No abnormal extra-axial collections are seen. No acute infarct is present. Intracranial arterial flow-voids are normal.
    \end{tcolorbox}
};

\node[anchor=north] (arrow2) at ([yshift=0.1cm]box2.south) {
    \huge $\downarrow$
};

\node[anchor=north, text width=2\textwidth] (box3) at ([yshift=-0.4cm]box2.south) {
    \begin{tcolorbox}[title=Structured Report, width=\textwidth, colframe=blue!30!black, colback=blue!15, fonttitle=\bfseries]
    \begin{itemize}[leftmargin=*]
        \item Evidence of prior surgery? Yes; Kind of surgery performed? left parietal craniotomy
        \item Enhancing lesions? Yes; Biggest lesion length? 1 to 2cm; Side with more enhancing lesions? Left; Locations with enhancing lesions? Left frontal lobe; Number of enhancing lesions? One
        \item Non-enhancing lesions? No
        \item Herniation/midline shift? No
        \item Hydrocephalus? No; Type of hydrocephalus? NA
        \item White matter changes? No
        \item Pituitary gland normal? Yes
        \item Sella/parasellar abnormalities? No
        \item Evidence of invasion/compression? Nowhere
    \end{itemize}
    \end{tcolorbox}
};

\end{tikzpicture}
\caption[Example \#1 of processed report]{An example of our automated report processing. The re-writing correctly rephrases the sentences, making references to changes in findings and removes references to, e.g. kTrans findings. The structured information extraction correctly answers all our instructions.}
\label{fig:brain-gpt-proc-ex1}
\end{figure*}
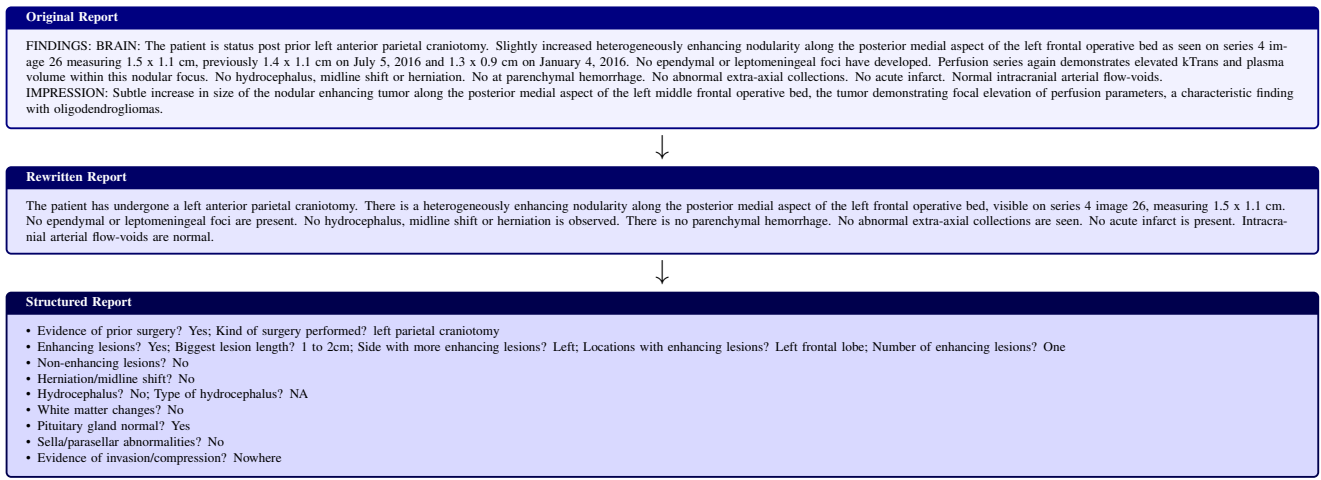

Annotating all the reports costs approximately \$1,600, which is significantly lower than the cost of expert annotation. Figure~\ref{fig:brain-prompt1} shows the report re-writing prompt and Figure~\ref{fig:brain-prompt2} shows the information extraction prompt. Example of this two-step processing are shown in Figures~\ref{fig:brain-gpt-proc-ex1},~\ref{fig:brain-gpt-proc-ex2}, and~\ref{fig:brain-gpt-proc-ex3}.

As there were around 80,000 long radiology reports, we used Python’s asyncio framework to process multiple reports in parallel with GPT-4. Each report underwent two API calls: one for re-writing and one for extracting structured information. We managed the API rate limits by processing reports in batches and including sleep time between batches. We also added a logit bias to avoid certain temporal medical terms (e.g., ``increase'', ``new'') in the answer. 
The temperature was set to 0.0 and top\_p to 1.0 for deterministic outputs. 
With parallelization, processing 80,000 reports took around 48 hours. GPT-4 performed significantly better than GPT-3.5 during preliminary comparisons.

\section{Motivation for DPPs}

Figure~\ref{fig:dpp-moti1} and~\ref{fig:dpp-moti2} in the Appendix illustrate how DPPs promote a more desirable feature diversity than pairwise repulsion. 

\begin{figure}[h]
  \centering
   \includegraphics[width=1\linewidth]{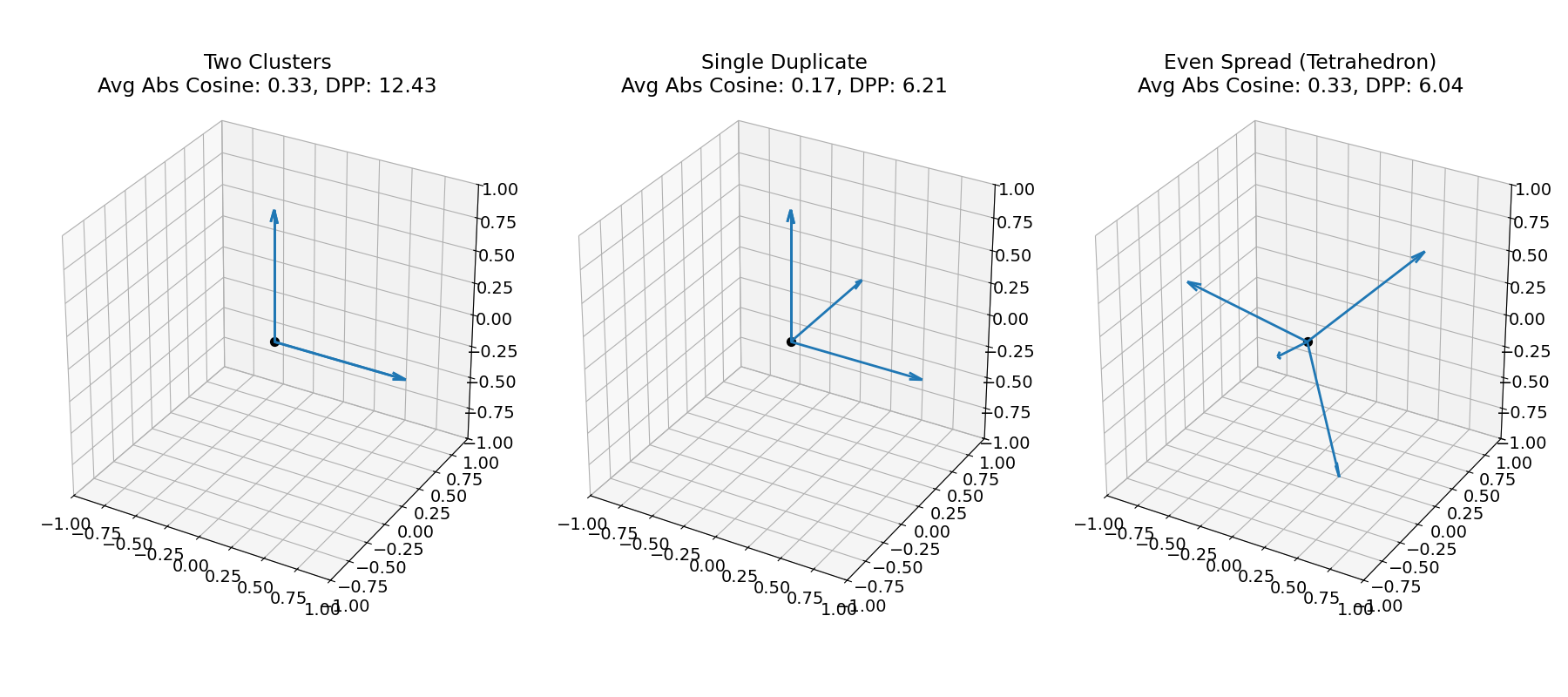}
   \caption{Visualization of the volume spanned by four 3D vectors under different configurations. In the left figure, both vectors are duplicates. In the middle one, one vector is a duplicate. The average pairwise cosine dissimilarity is the same for the left and right figures, even though the right figure represents a much more desirable spread of the vectors. The DPP is much lower for that figure thus minimizing the DPP is superior.}
   \label{fig:dpp-moti1}
\end{figure}

\begin{figure}[h]
  \centering
   \includegraphics[width=1\linewidth]{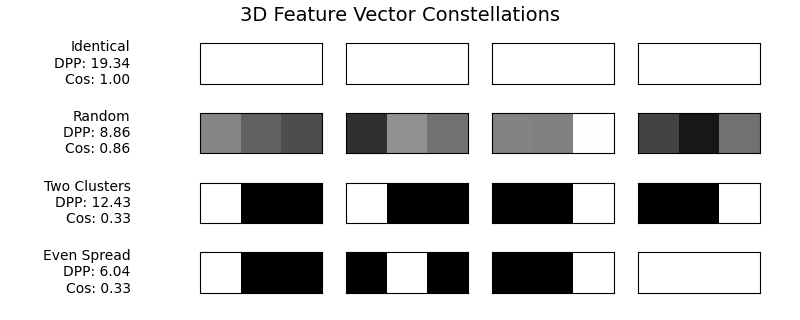}
   \caption{Visualization of four 3D vectors under different configurations, representing attention maps over a $1 \times 3$ image. It shows that pairwise cosine dissimilarity (``Cos'') equally rewards an even spread of attention maps (desired) and the embeddings grouping into two clusters (not desired). In contrast, the DPP rewards the even spread the most.}
   \label{fig:dpp-moti2}
\end{figure}

\section{Implementation and Training Details} \label{app:trai9}

In this section, we discuss the implementation details of our pre-training and downstream evaluation. All code and models will be made public.

\subsection{Pre-Training}

Training parameters were determined empirically, with the final set provided in Table~\ref{impl:pretrain}. Contrary to the general assumption that big batches lead to improved performance for image-text contrastive learning, our results consistently improved for comparitvely small batch sizes, in the range of 25-32. This allowed us to train each model on a single A/H100 GPU. Our experiments also found that the Q-Former's language modeling loss consistently improved performance across nearly all configurations, while image-text matching did not yield benefits, leading us to omit the image-text matching loss. We also found that using a biomedically pre-trained BERT outperformed the standard BERT pre-training version in all evaluated scenarios. All model weights were selected based on the best average metrics on the development set. For \dset, all models were trained with the same image processing: $1mm\times1mm\times1mm$ voxel spacing, intensity normalization, and resizing to $32\times256\times256$. Preliminary analysis on \dset \ also showed lowered performance with standard data augmentation such as Gaussian noise, image rotation and translation, as well as random view cropping, and was removed from subsequent analyses. We also explored various text augmentation techniques. In ``text dropout'', a given share of text tokens is randomly masked during the forward pass. In ``sentence dropout'', different sentences, i.e., clinical features, are randomly removed from the reports. In ``PVA dropout,'' we randomly drop a given share of paired multi-view embeddings and clinical features from the overall similarity matrix. None of these techniques have led to performance gains.

\subsection{Downstream Tasks}

We evaluated our pre-training methods by fine-tuning on several downstream tasks. When feasible, our hyperparameters were selected via grid search. The ADNI hyperparameter are given in Table~\ref{impl:cls}. For ADNI, image preprocessing was performed using Clinica's t1-volume-tissue-segmentation pipeline. For report generation, we follow the parameters chooses in the Llama paper~\cite{grattafiori2024llama}. We used a batch size of 1024 and learning rate of 0.0002. We use an AdamW optimizer with a cosine decay and a warm-up ratio of 0.3. For segmentation, we used nnUNet as the baseline model, fine-tuning it with an initial learning rate of 1e-2 and a weight decay of 3e-5. The training pipeline included standard nnUNet pre-processing, data augmentation was not used. Result model weights were selected based on the highest mean Dice score for BraTS-2021 and the best Lesion-Wise metrics for BraTS-2023-METS on the validation set. All model weights were selected based on the best average metrics on the development set.

\begin{table}[h]
  \centering
  \resizebox{\columnwidth}{!}{%
  \begin{tabular}{@{}lccc@{}}
    \toprule
    Parameter & 1\% & 10\% & 100\% \\
    \midrule
    Batch Size & 16 & 32 & 32 \\
    Learning Rate & 1.00E-06 & 1.00E-05* & 1.00E-05* \\
    Training Precision & \multicolumn{3}{c}{Bfloat16} \\
    Augmentation & Yes & No & No \\
    Trained Layers & MLP Only & All & All \\
    MLP size & \multicolumn{3}{c}{2 layers} \\
    \bottomrule
  \end{tabular}%
  }
  \caption{Implementation details of our Alzheimer classification downstream task. *For ViT we used 1.00E-06 across all data amounts. Augmentation consisted of: random flipping, random intensity scaling, random intensity shifting, adding gaussian noise, gaussian smoothing, random contrast adjustment, and random low resolution simulation. More details can be found in our code.}
  \label{impl:cls}
\end{table}

\begin{table}[h!]
  \centering
  \resizebox{\columnwidth}{!}{%
  \begin{tabular}{@{}lcc@{}}
    \toprule
    Parameter & all models on \dset & \model on \bim \\
    \midrule
    Batch Size & 32 & 25 \\
    3D Vision Model $M$ & Densenet-121/ViT/ResNet-50 & Densenet-121\\
    Weights Init. of $M$ & \multicolumn{2}{c}{None} \\
    Architecture of $E_{I/R}$ & \multicolumn{2}{c}{BERT-base\footnote{188M parameters, cross-attention layers randomly initialized}} \\
    Weights Init. of $E_{I/R}$ & \multicolumn{2}{c}{BiomedBERT\footnote{microsoft/BiomedNLP-BiomedBERT-base-uncased-abstract-fulltext}} \\
    Learning Rate $M$ & \multicolumn{2}{c}{5.00E-04*} \\
    Learning Rate $E_{I/R}$ & \multicolumn{2}{c}{5.00E-05} \\
    Max. Text Length $E_R$ & \multicolumn{2}{c}{256} \\
    $N_Q$ (\# of Query Tokens) & \multicolumn{2}{c}{32} \\
    Cross-Attention Frequency & \multicolumn{2}{c}{2} \\
    Max. Number of Sentences & \multicolumn{2}{c}{20} \\
    Training Precision & \multicolumn{2}{c}{Bfloat16} \\
    Augmentation & \multicolumn{2}{c}{None} \\
    \bottomrule
  \end{tabular}%
  }
  \caption{Implementation details of our pre-training. Except for \bim, the batch size was chosen to be maximal given compute resources. *For ViT, we used a lower learning rate of 1.00E-07.}
  \label{impl:pretrain}
\end{table}

\subsection{BIMCV-R Dataset} \label{app:bim}

\begin{figure*}[h]
\centering
  \includegraphics[width=1\linewidth]{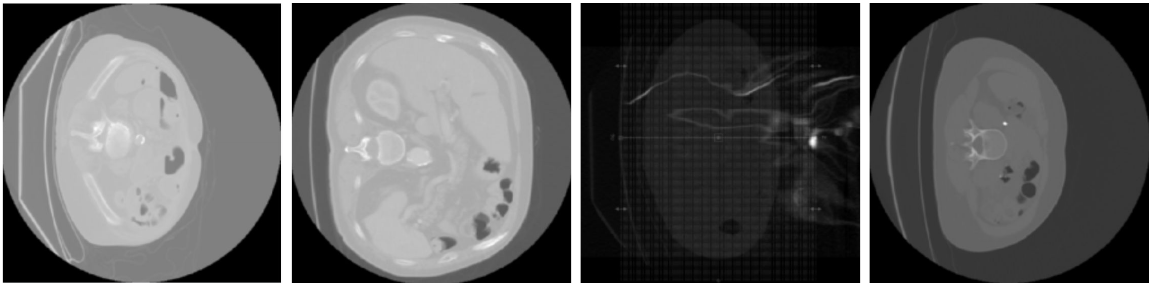}
  \caption{\bim example images of localizer scans or where the middle slice is already in the abdomen or pelvis.}
  \label{fig:bimcvr}
\end{figure*}

We found quality issues with the \bim dataset that may explain the overall lower performance obtained on this dataset compared to \dset. Figure~\ref{fig:bimcvr} shows how for some images the middle slice (depicted) is already no longer in the lung, suggesting that the scan mainly depicts other body parts. Several images also seem to depict localizer scans, which makes it difficult to connect them to radiology reports. Appropriate processing of these images would likely lead to significant performance improvements. 

\section{Additional Results}

In this section, we provide more detailed results and examples.

\subsection{Alzheimer's Classification}

More detailed results for Alzheimer's classification are provided in Table~\ref{tab:adni_results}.

\begin{table*}[h!]
  \centering
  \resizebox{\textwidth}{!}{%
  \begin{tabular}{llcccccccccccc}
    \toprule
    \multicolumn{2}{c}{Pre-training Approaches} & \multicolumn{4}{c}{1\% Training Data (n=19)} & \multicolumn{4}{c}{10\% Training Data (n=193)} & \multicolumn{4}{c}{100\% Training Data (n=1,932)} \\ 
    \cmidrule(lr){3-6} \cmidrule(lr){7-10} \cmidrule(lr){11-14}
    Vision Model \(M\) & Weight Init. & Alz. & Normal & MCI & \(\mu\) & Alz. & Normal & MCI & \(\mu\) & Alz. & Normal & MCI & \(\mu\) \\
    \midrule
    Densenet-121
      & Random   & 0.523   & 0.513   & 0.527   & 0.521 [0.495, 0.547] 
                 & 0.640   & 0.560   & 0.498   & 0.567 [0.535, 0.596]
                 & 0.724   & 0.629   & 0.535   & 0.629 [0.608, 0.649] \\
      & CLS      & 0.514   & 0.517   & 0.511   & 0.514 [0.487, 0.538]
                 & 0.614   & 0.598   & 0.523   & 0.578 [0.555, 0.602]
                 & 0.720   & 0.628   & 0.556   & 0.635 [0.612, 0.650] \\
      & Q-Former & 0.565   & 0.525   & 0.486   & 0.526 [0.506, 0.547]
                 & 0.688   & 0.627   & 0.550   & 0.623 [0.604, 0.640]
                 & 0.747   & 0.662   & 0.581   & 0.663 [0.651, 0.681] \\
      & \model   & 0.560   & 0.559   & 0.505   & 0.543 [0.497, 0.579]
                 & 0.720   & 0.644   & 0.518   & 0.628 [0.606, 0.653]
                 & 0.793   & 0.687   & 0.505   & 0.661 [0.650, 0.672] \\
    \midrule[\lightrulewidth]
    ResNet-50
      & Random   & 0.497   & 0.566   & 0.541   & 0.535 [0.497, 0.569]
                 & 0.516   & 0.529   & 0.541   & 0.530 [0.498, 0.561]
                 & 0.590   & 0.525   & 0.528   & 0.548 [0.514, 0.586] \\
      & \model   & 0.527   & 0.531   & 0.532   & 0.530 [0.500, 0.556]
                 & 0.621   & 0.456   & 0.452   & 0.510 [0.490, 0.532]
                 & 0.636   & 0.542   & 0.533   & 0.569 [0.519, 0.612] \\
    \midrule[\lightrulewidth]
    ViT
      & Random   & 0.517   & 0.485   & 0.473   & 0.492 [0.471, 0.512]
                 & 0.554   & 0.491   & 0.502   & 0.515 [0.495, 0.532]
                 & 0.528   & 0.473   & 0.515   & 0.505 [0.476, 0.533] \\
      & \model   & 0.518   & 0.498   & 0.460   & 0.491 [0.458, 0.523]
                 & 0.607   & 0.555   & 0.467   & 0.543 [0.522, 0.561]
                 & 0.622   & 0.521   & 0.450   & 0.531 [0.513, 0.551] \\
    \bottomrule
  \end{tabular}%
  }
  \caption{Evaluation results (AUC scores) for different initialisations using 1\%, 10\%, and 100\% of training data. “Alz.” stands for Alzheimer’s disease and “MCI” for mild cognitive impairment. The column \(\mu\) is the average of the per-class AUC scores computed on the balanced test set; only this column displays the confidence interval.}
  \label{tab:adni_results}
\end{table*}

\subsection{Report Generation}

Results in Table~\ref{tab:rgen} show that, for the GREEN metric, we see sizeable gains from general vision-language pre-training but no big difference between traditional Q-Former training versus the \model framework. The GREEN metric provides a structured clinical assessment by identifying key radiology report errors derived from expert evaluations. However, it was developed mainly on chest X-ray reports, and therefore, its applicability to out-of-distribution modalities is limited. The authors evaluated their metric on an abdomen CT dataset and found a high absolute error (5.31). I provide this metric as it can offer a rough assessment of the clinical correctness, however, it is likely not well suited to assess minor performance differences in brain MRI reports. An inspection of the generated evaluations by GREEN confirms that they contain many errors. A more dependable approach would involve direct human evaluation or leveraging a stronger LLM such as GPT-4 for assessment.

\begin{figure*}[h]
\centering
  \includegraphics[width=1\linewidth]{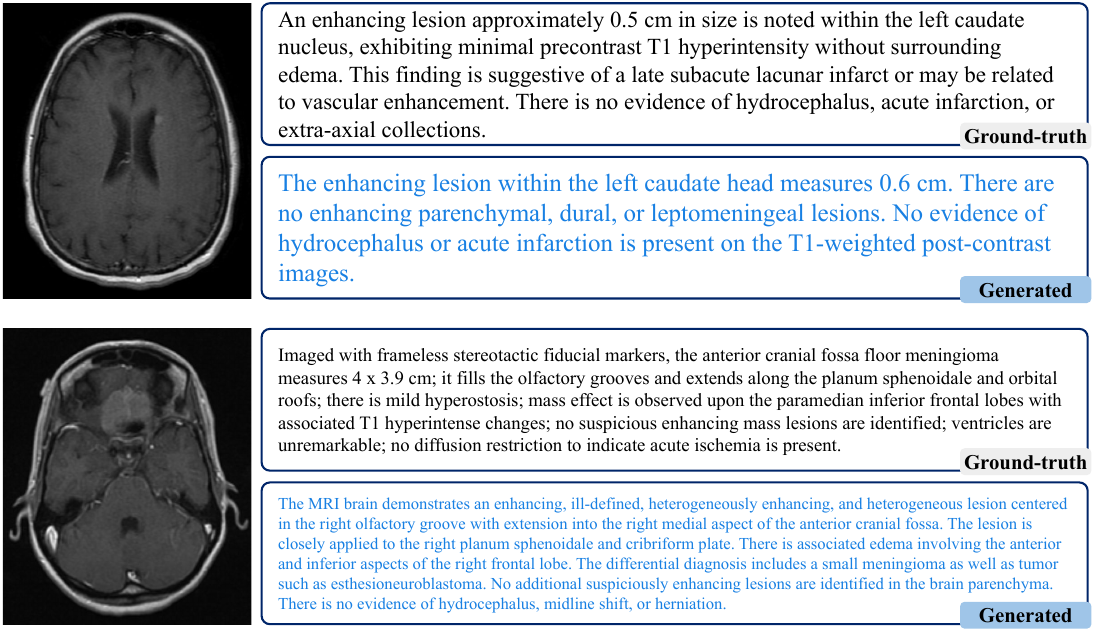}
    \caption{Two example MRI reports generated by our VLM with a \model backbone. The reports largely capture the correct image findings.}
    \label{fig:gen-example}
\end{figure*}

Figure~\ref{fig:gen-example} shows examples of generated reports.

\subsection{Segmentation}

Figure~\ref{tab:brats2021_results} and Figure~\ref{tab:brats2023_results} provide more detailed results on segmentation.

\begin{table*}[h!]
  \centering
  \resizebox{\textwidth}{!}{%
  \begin{tabular}{llccccccccc}
    \toprule
    \multicolumn{2}{c}{Pre-training Approaches} & \multicolumn{3}{c}{1\% Training Data (n=12)} & \multicolumn{3}{c}{10\% Training Data (n=120)} & \multicolumn{3}{c}{100\% Training Data (n=1200)} \\ 
    \cmidrule(lr){3-5} \cmidrule(lr){6-8} \cmidrule(lr){9-11}
    Vision Model \(M\) & Weight Init. & Whole Tumor & Tumor Core & Enhanced Tumor & Whole Tumor & Tumor Core & Enhanced Tumor & Whole Tumor & Tumor Core & Enhanced Tumor \\
    \midrule
    Densenet-121
      & Random   & 0.780 & 0.646 & 0.585 
                 & 0.875 & 0.791 & 0.710 
                 & 0.903 & 0.865 & 0.779 \\
    \midrule
    Densenet-121
      & \model         & 0.796 & 0.633 & 0.580 
                 & 0.870 & 0.785 & 0.707 
                 & 0.903 & 0.864 & 0.776 \\
    \bottomrule
  \end{tabular}%
  }
  \caption{Segmentation performance (Dice scores) for different pre-training initialisations using 1\%, 10\%, and 100\% of the training data. The values correspond to the Dice scores for the Whole Tumor, Tumor Core, and Enhanced Tumor regions.}
  \label{tab:brats2021_results}
\end{table*}

\begin{table*}[h!]
  \centering
  \resizebox{\textwidth}{!}{%
  \begin{tabular}{llccccccccc}
    \toprule
    \multicolumn{2}{c}{Pre-training Approaches} & \multicolumn{3}{c}{1\% Training Data (n=12)} & \multicolumn{3}{c}{10\% Training Data (n=120)} & \multicolumn{3}{c}{100\% Training Data (n=1200)} \\ 
    \cmidrule(lr){3-5} \cmidrule(lr){6-8} \cmidrule(lr){9-11}
    Vision Model \(M\) & Weight Init. & Whole Tumor & Tumor Core & Enhanced Tumor & Whole Tumor & Tumor Core & Enhanced Tumor & Whole Tumor & Tumor Core & Enhanced Tumor \\
    \midrule
    Densenet-121
      & Random   & 0.780 & 0.646 & 0.585 
                 & 0.875 & 0.791 & 0.710 
                 & 0.922 & 0.854 & 0.761 \\
    \midrule
    \model
      & \model         & 0.796 & 0.633 & 0.580 
                 & 0.870 & 0.785 & 0.707 
                 & 0.925 & 0.867 & 0.762 \\
    \bottomrule
  \end{tabular}%
  }
  \caption{Segmentation performance (Lesion-wise Dice scores) for different pre-training initialisations using 1\%, 10\%, and 100\% of the training data. The values correspond to the lesion-wise Dice scores for the Whole Tumor, Tumor Core, and Enhanced Tumor regions.}
  \label{tab:brats2023_results}
\end{table*}
\begin{figure}[htb]
    \small
    \begin{tcolorbox}[
        colback=gray!5,    
        boxrule=0.5pt,     
        arc=0pt            
    ]
    You are a highly experienced radiologist. Accurately answer the questions below based on the given brain MRI report. Your output must be in json format.
    \\ \\
    (a) For each question, choose the appropriate answer (wording must match exactly). If answers are mutually exclusive, choose one. If multiple answers can apply, list all that are true, separated by semicolons (";"). \\
    (b) If the MRI report does not contain information to answer a specific question, use the default answer indicating a normal status. \\
    (c) Note the following assumptions: meningiomas are considered enhancing lesions; burr holes and ventriculostomy and Ommaya catheters are considered prior surgery; punctate lesions are less than 1cm.
    \\ \\
    Questions (Answer options): \\
    Is there evidence of prior surgery? (Yes / No) \\
    What kind of surgery was performed? (NA / left frontal craniotomy; right frontal craniotomy; left parietal craniotomy; right parietal craniotomy; left temporal or pterional craniotomy; right temporal or pterional craniotomy ; left occipital craniotomy; right occipital craniotomy) \\
    Are there any enhancing lesions? (Yes / No) \\
    What is the length of the biggest mass lesion? (NA / Less than 1cm / 1 to 2cm / More than 2cm) \\
    Which side of the brain has more enhancing lesions? (NA / Left / Right) \\
    List all the locations that contain enhancing lesions. (NA / Left frontal lobe; Right frontal lobe; Left parietal lobe; Right parietal lobe; Left temporal lobe; Right temporal lobe; Left occipital lobe; Right occipital lobe; Left thalamus or basal ganglia; Right thalamus or basal ganglia; Cerebellum; brainstem; cervical spinal cord) \\
    How many enhancing lesions are there? (NA / One / Between 2 and 6 / Between 7 and 15 / More than 15) \\
    Is there a herniation or midline shift? (Yes / No) \\
    Are there any signs of white matter disease (e.g., leukoaraiosis or leukoencephalopathy)? (Yes / No) \\
    Is the pituitary gland normal in appearance? (Yes / No) \\
    Are there abnormalities in the sella or parasellar regions? (Yes / No) \\
    Where is there evidence of invasion into or compression of adjacent structures? (Nowhere / ventricles; brainstem; cranial nerves) \\
    \\
    Brain MRI report: \$\{Insert processed report\}\$
    \end{tcolorbox}
    \caption[Report information extraction prompt]{The final prompt that was used to extract structured information from the reports. Note that the structured information is mainly used to gain more understanding of our dataset and evaluate models. The only model in this paper trained on these labels is the Classification baseline which we compare against in Table~\ref{tab:rgen} and Figure~\ref{fig:adni}.}
    \label{fig:brain-prompt2}
\end{figure}

\begin{figure*}[h!]
\centering
\begin{tikzpicture}[remember picture, scale=0.5, transform shape]

\node[anchor=north, text width=2\textwidth] (box1) at (0, 0) {
    \begin{tcolorbox}[title=Original Report, width=\textwidth, colframe=blue!50!black, colback=blue!5, fonttitle=\bfseries]
    FINDINGS: BRAIN: Unchanged plaque-like left parietal convexity enhancing meningioma at 0.9 x 0.3 cm with mild overlying hyperostosis; No suspicious parenchymal enhancing mass lesions, increase in Ktrans or increase in plasma volume; Normal ventricles; no diffusion restriction to indicate acute ischemia.
    \\
    IMPRESSION: Since February 2015, no brain metastasis. Unchanged subcm left parietal convexity meningioma. Unchanged mild enlargement of the pituitary gland. Unchanged benign left orbital lesion, possibly a hemangioma or dermoid/epidermoid.
    \end{tcolorbox}
};

\node[anchor=north] (arrow1) at ([yshift=0.1cm]box1.south) {
    \huge $\downarrow$
};

\node[anchor=north, text width=2\textwidth] (box2) at ([yshift=-0.4cm]box1.south) {
    \begin{tcolorbox}[title=Rewritten Report, width=\textwidth, colframe=blue!40!black, colback=blue!10, fonttitle=\bfseries]
    Plaque-like left parietal convexity enhancing meningioma at 0.9 x 0.3 cm with mild overlying hyperostosis is present. No suspicious parenchymal enhancing mass lesions are observed. Ventricles appear normal. There is no diffusion restriction indicating acute ischemia.
    \end{tcolorbox}
};

\node[anchor=north] (arrow2) at ([yshift=0.1cm]box2.south) {
    \huge $\downarrow$
};

\node[anchor=north, text width=2\textwidth] (box3) at ([yshift=-0.4cm]box2.south) {
    \begin{tcolorbox}[title=Structured Report, width=\textwidth, colframe=blue!30!black, colback=blue!15, fonttitle=\bfseries]
    \begin{itemize}[leftmargin=*]
        \item Evidence of prior surgery? No
        \item Kind of surgery performed? NA
        \item Enhancing lesions? Yes
        \item Non-enhancing lesions? No
        \item Biggest lesion length? <1cm
        \item Side with more enhancing lesions? Left
        \item Locations with enhancing lesions? Left parietal lobe
        \item Number of enhancing lesions? One
        \item Herniation/midline shift? No
        \item Hydrocephalus? No
        \item Type of hydrocephalus? NA
        \item White matter changes? No
        \item Grey matter abnormalities? No
        \item Pituitary gland normal? Yes
        \item Sella/parasellar abnormalities? No
        \item Evidence of invasion/compression? Nowhere
    \end{itemize}
    \end{tcolorbox}
};

\end{tikzpicture}
\caption[Example \#2 of processed report]{An example of our automated report processing. Here, the intermediate re-written report omits the enlargement of the pituitary gland and left orbital lesion.}
\label{fig:brain-gpt-proc-ex2}
\end{figure*}
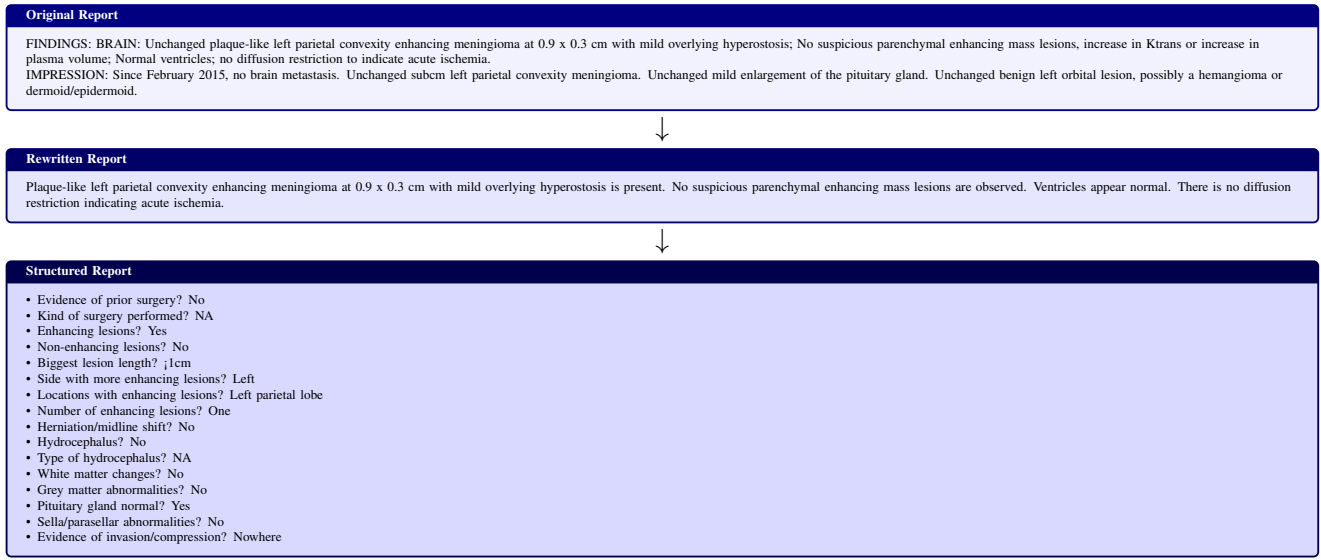

\begin{figure*}[h!]
\centering
\begin{tikzpicture}[remember picture, scale=0.5, transform shape]

\node[anchor=north, text width=2\textwidth] (box1) at (0, 0) {
    \begin{tcolorbox}[title=Original Report, width=\textwidth, colframe=blue!50!black, colback=blue!5, fonttitle=\bfseries]
    FINDINGS: BRAIN: The peripherally enhancing centrally cystic/necrotic metastasis within inferior right frontal lobe bulging into frontal horn of the right lateral ventricle has not significantly changed in size when allowing for differences in image acquisition. Extensive right anterior hemispheric vasogenic edema which extends into the basal ganglia and anterior thalamus appears mildly improved. Mass effect resulting in right-to-left midline shift has mildly improved as well. The Ommaya catheter remains in place with its tip within necrotic core of the right inferior frontal lobe mass. The nodular dural based thickening along the anteroinferior margin of the right frontal burr hole has increased now measuring 1.8 x 1.3 cm on the axial images, having previously measured 1.1 x 0.8 cm. On the coronal images, it measures 1.5 x 0.8 cm and likely represent focal dural based tumor. The previously seen superficial focus of enhancement along the right inferior frontal gyrus appears less conspicuous. Small enhancing lesion adjacent to upper aspect of the Ommaya catheter has increased in size now measuring 4 to 5 mm. Right high parafalcine parietal hemorrhagic lesion has not significantly changed. Background white matter changes are stable. \\
    IMPRESSION: Compared to December 21, 2011; 1. Mild progression of dural based disease along the anteroinferior margin of the right frontal burr hole. 2. Mild progression of the metastasis adjacent to upper aspect of the right Ommaya catheter. 3. Stable hemorrhagic lesion within high right parafalcine parietal lobe. 4. The heterogeneous right inferior frontal lobe mass its stable in size however, right hemispheric edema and mass effect have mildly improved.
    \end{tcolorbox}
};

\node[anchor=north] (arrow1) at ([yshift=0.1cm]box1.south) {
    \huge $\downarrow$
};

\node[anchor=north, text width=2\textwidth] (box2) at ([yshift=-0.4cm]box1.south) {
    \begin{tcolorbox}[title=Rewritten Report, width=\textwidth, colframe=blue!40!black, colback=blue!10, fonttitle=\bfseries]
    The peripherally enhancing centrally cystic/necotic metastasis is located within the inferior right frontal lobe, bulging into the frontal horn of the right lateral ventricle. There is extensive right anterior hemispheric vasogenic edema extending into the basal ganglia and anterior thalamus. Mass effect results in right-to-left midline shift. The Ommaya catheter is in place with its tip within the necrotic core of the right inferior frontal lobe mass. There is nodular dural based thickening along the anteroinferior margin of the right frontal burr hole, measuring 1.8 x 1.3 cm on the axial images. On the coronal images, it measures 1.5 x 0.8 cm and likely represents a focal dural based tumor. There is a superficial focus of enhancement along the right inferior frontal gyrus. A small enhancing lesion is adjacent to the upper aspect of the Ommaya catheter, measuring 4 to 5 mm. There is a right high parafalcine parietal hemorrhagic lesion. Background white matter changes are present.
    \end{tcolorbox}
};

\node[anchor=north] (arrow2) at ([yshift=0.1cm]box2.south) {
    \huge $\downarrow$
};

\node[anchor=north, text width=2\textwidth] (box3) at ([yshift=-0.4cm]box2.south) {
    \begin{tcolorbox}[title=Structured Report, width=\textwidth, colframe=blue!30!black, colback=blue!15, fonttitle=\bfseries]
    \begin{itemize}[leftmargin=*]
        \item Is there evidence of prior surgery? Yes
        \item What kind of surgery was performed? Right frontal craniotomy
        \item Are there any enhancing lesions? Yes
        \item Are there any non-enhancing lesions? Yes
        \item What is the length of the biggest lesion? 1 to 2 cm
        \item Which side of the brain has more enhancing lesions? Right
        \item List all the locations that contain enhancing lesions. Right frontal lobe; Right thalamus or basal ganglia
        \item How many enhancing lesions are there? Between 2 and 6
        \item Is there a herniation or midline shift? Yes
        \item Is there hydrocephalus? No
        \item What kind of hydrocephalus is there? NA
        \item Are there white matter changes suggestive of other pathologies? Yes
        \item Are there abnormalities in the cortical or subcortical grey matter suggestive of other pathologies? No
        \item Is the pituitary gland normal in appearance? Yes
        \item Are there abnormalities in the sella or parasellar regions? No
        \item Where is there evidence of invasion into or compression of adjacent structures? Ventricles
    \end{itemize}
    \end{tcolorbox}
};

\end{tikzpicture}
\caption[Example \#3 of processed report]{An example of a long and complex report that was processed successfully.}
\label{fig:brain-gpt-proc-ex3}
\end{figure*}
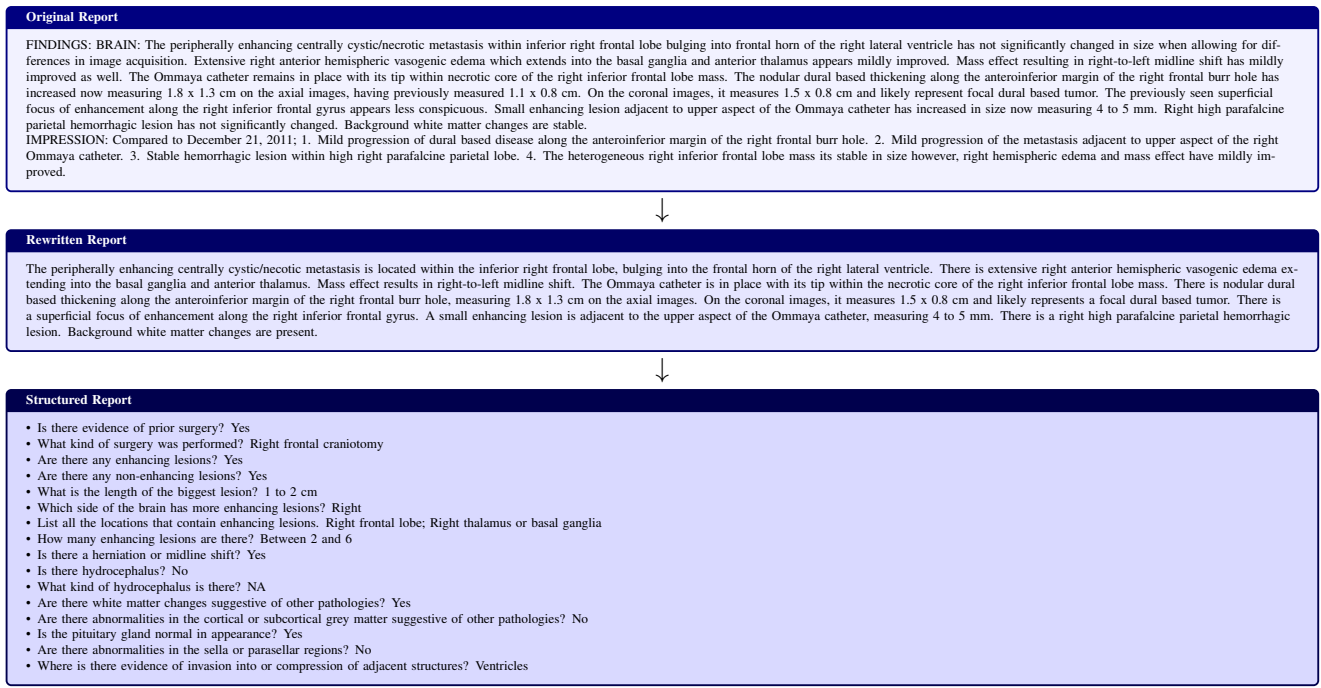

\end{document}